\documentclass{article}

\usepackage{microtype}
\usepackage{graphicx}
\usepackage{subfigure}
\usepackage{booktabs} 
\usepackage{hyperref}

\usepackage[accepted]{icml2019}

\usepackage{xspace}
\usepackage{dblfloatfix}
\newcommand{\imgep}{\textsc{imgep}\xspace}
\newcommand{\uvfa}{\textsc{uvfa}\xspace}

\newcommand{\her}{\textsc{her}\xspace}
\newcommand{\unicorn}{\textsc{unicorn}\xspace}
\newcommand{\saggriac}{\textsc{sagg-riac}\xspace}
\newcommand{\ddpg}{\textsc{ddpg}\xspace}

\newcommand{\sacx}{\textsc{sac-x}\xspace}
\newcommand{\te}{\textsc{mg-me}\xspace}

\newcommand{\curious}{\textsc{curious}\xspace}

\newcommand{\muvfa}{\textsc{m-uvfa}\xspace}

\newcommand{\tdthree}{\textsc{td3}\xspace}

\definecolor{myred}{rgb}{0.8,0,0}

\usepackage{wrapfig}
\usepackage{float}
\usepackage{amssymb}
\usepackage{amsmath}

\newcommand\fauxsc[1]{\fauxschelper#1 \relax\relax}
\def\fauxschelper#1 #2\relax{%
  \fauxschelphelp#1\relax\relax%
  \if\relax#2\relax\else\ \fauxschelper#2\relax\fi%
}
\def\Hscale{.85}\def\Vscale{.74}\def\Cscale{1.12}
\def\fauxschelphelp#1#2\relax{%
  \ifnum`#1>``\ifnum`#1<`\{\scalebox{\Hscale}[\Vscale]{\uppercase{#1}}\else%
    \scalebox{\Cscale}[1]{#1}\fi\else\scalebox{\Cscale}[1]{#1}\fi%
  \ifx\relax#2\relax\else\fauxschelphelp#2\relax\fi}

\icmltitlerunning{CURIOUS: Intrinsically Motivated Modular Multi-Goal Reinforcement Learning}

\begin{document}
\twocolumn[
\icmltitle{CURIOUS: Intrinsically Motivated Modular \\ Multi-Goal Reinforcement Learning}


\begin{icmlauthorlist}
\icmlauthor{C\'edric Colas}{in}
\icmlauthor{Pierre Fournier}{up}
\icmlauthor{Olivier Sigaud}{up}
\icmlauthor{Mohamed Chetouani}{up}
\icmlauthor{Pierre-Yves Oudeyer}{in}
\end{icmlauthorlist}

\icmlaffiliation{in}{Flowers Team, Inria and Ensta ParisTech, FR.}
\icmlaffiliation{up}{ISIR, Sorbonne Univ., Paris, FR.}

\icmlcorrespondingauthor{C\'edric Colas}{cedric.colas@inria.fr}

\icmlkeywords{Machine Learning, ICML}

\vskip 0.3in
]



\printAffiliationsAndNotice{}  
\begin{abstract}

In open-ended environments, autonomous learning agents must set their own goals and build their own curriculum through an intrinsically motivated exploration. They may consider a large diversity of goals, aiming to discover what is controllable in their environments, and what is not. Because some goals might prove easy and some impossible, agents must actively select which goal to practice at any moment, to maximize their overall mastery on the set of learnable goals. This paper proposes \curious, an algorithm that leverages 1) a modular Universal Value Function Approximator with hindsight learning to achieve a diversity of goals of different kinds within a unique policy and 2) an automated curriculum learning mechanism that biases the attention of the agent towards goals maximizing the absolute learning progress. Agents focus sequentially on goals of increasing complexity, and focus back on goals that are being forgotten. Experiments conducted in a new modular-goal robotic environment show the resulting developmental self-organization of a learning curriculum, and demonstrate properties of robustness to distracting goals, forgetting and changes in body properties.

\end{abstract}

\section{Introduction}

    \begin{figure}[ht]
    \centering
        \includegraphics[width=0.6\columnwidth]{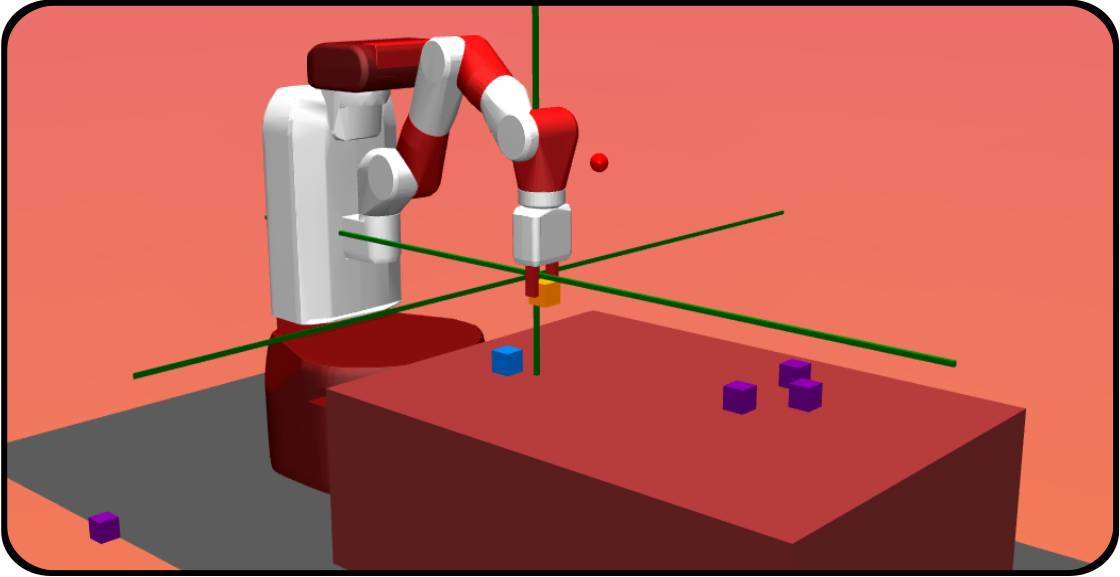}
         \caption{\textbf{The \emph{Modular Goal Fetch Arm} environment.} An intrinsically motivated agent can set its own (modular) goals (Reach, Push, Pick and Place, Stack), with multiple objects and distractors.}  
    \label{fig:mod_fetch}
    \end{figure}
    
    
    In \emph{autonomous continual learning}, agents aim to discover repertoires of skills in an ever-changing open-ended world, and without external rewards. In such realistic environments, the agent must be endowed with intrinsic motivations to explore the diversity of ways in which it can control its environment. One important form of intrinsic motivation system is the ability to autonomously set one's own goals and self-organize one's own curriculum. This challenge can be tackled within the framework of Intrinsically Motivated Goal Exploration Processes (\imgep) \cite{baranes2013active,forestier2017intrinsically}, leveraging computational models of autonomous development in human infants.
            
    \paragraph{Modular goal representation.}In a same environment, an agent might want to `put the cube in position $x$' or to `reach position $y$' for any $x$ or $y$. Here, describing the full goal space requires modular goal representations. Goals are organized by \textit{modules}, where \textit{module} refers to the pair of a reward function and a goal space $M_i=(R_{M_i, g_i \in \mathcal{G}_{\mathcal{M}_i}}, \mathcal{G}_{\mathcal{M}_i})$. The reward function describes a set of constraints that must be satisfied by the agent's state (e.g. Reach), given a continuous parameter (e.g. $g_i=y$) evolving in the associated goal space (e.g. 3D Euclidean space), see Fig.~\ref{fig:mod_fetch}. 
    
    While flat multi-goal problems with continuous \citep{schaul2015universal, andrychowicz2017hindsight, plappert2018multi} or discrete goals \citep{mankowitz2018unicorn, riedmiller2018learning} have been explored in the past, only few works tackle the problem of modular multi-goal learning \citep{forestier2016modular}, none in an RL setting. Here, we present \curious\footnote{ \curious stands for Continual Universal Reinforcement learning with Intrinsically mOtivated sUbstitutionS.}, a modular multi-goal reinforcement learning (RL) algorithm that uses intrinsic motivations to efficiently learn a continuous set of diverse goals using modular goal representations. To build an algorithm able to learn modular goals, one must answer the following questions: 1) How to choose the action policy architecture? 2) How to select the next module and goal to practice and learn about? 3) How to efficiently transfer knowledge between modules and goals?

            
    \begin{table*}[bh!]
        \caption{\textbf{Classification of multi-goal approaches}. \underline{Underlined}: Algorithms internally generating goals (\imgep), (*) using LP-based intrinsic motivations. \emph{Italic}: Population-based algorithms (non-RL). \textbf{Bold}: Algorithms proposed in this paper.}
        \label{sample-table}
        \vskip 0.15in
        \begin{center}
        \begin{small}
        \begin{tabular}{lccr}
            \toprule
              & \textsc{$n$ goals, $n$ policies} & \textsc{$n$ goals, 1 policy} \\
            
            \midrule
               \textsc{Flat repr.} & \textsc{Goal-experts}: & \textsc{Multi-goal}: \uvfa \cite{schaul2015universal}\\
            & \cite{kaelbling1993learning} & \her \cite{andrychowicz2017hindsight}  \\
            &  \sacx \cite{riedmiller2018learning} & \unicorn \cite{mankowitz2018unicorn}\\
            & \underline{\textit{\fauxsc{sagg-riac}*}} \cite{baranes2013active}& \\
            \midrule
                \textsc{Modular repr.}      & \textsc{mod-goal-experts}: \underline{\textit{\fauxsc{macob}*}} \cite{forestier2016modular} & \textsc{modular-multi-goals}: \\
                & \textsc{multi-goal module-experts}:  \underline{\textbf{\te}}& \underline{\textbf{\muvfa}}, \underline{\textbf{\curious*}} \\

        \end{tabular}
        \end{small}
        \end{center}
        \vskip -0.1in
    \end{table*}


    \paragraph{Related work.}\citet{kaelbling1993learning} proposed the first algorithm able to leverage cross-goal learning to address a discrete set of goals. For each goal, the algorithm learned a specific value function using Q-learning (\emph{goal-experts} approach). More recently, \citet{schaul2015universal} proposed Universal Value Function Approximators (\uvfa), a unique policy able to address an infinity of goals by concatenating the current state and goal to feed both the policy and the value function. In \unicorn, \uvfa is used to address a discrete set of goals in parallel: reaching different objects in a visual world \citep{mankowitz2018unicorn}. \sacx implements multi-task RL where easy tasks are considered as auxiliary tasks to help learning about the hardest task (placing cubes inside a closed box) \citep{riedmiller2018learning}. Here, one network is trained for each task and the collected transitions are shared (\emph{goal-experts} approach). In other works from multi-task RL \citep{teh2017distral, espeholt2018impala, hessel2018multi}, agents do not represent explicitly the current task and aim at maximizing the overall reward. Finally, within the Intrinsically Motivated Goal Exploration Processes framework (\imgep), \citet{forestier2016modular} proposed \textsc{macob}, an algorithm able to target modular goals using a population-based algorithm that mutates and replays controllers experienced in the past. \textsc{macob} maintains a population of solutions, one for each goal (\emph{modular goal-experts} approach), see \citet{mai2012active} for a similar approach. This enables efficient cross-goal learning in high-dimensional goal spaces, but is limited by the memory-based representation of policies.

    Multi-goal approaches prove better than simply training a policy per goal because knowledge can be transferred between different goals using off-policy learning and hindsight learning \citep{andrychowicz2017hindsight}. Off-policy learning enables the use of any transition to improve the current policy: transitions collected from an older version of the current policy \citep{lillicrap2015continuous}, from a population of exploratory policies \citep{colas2018gep}, or even from demonstrations \citep{vevcerik2017leveraging}. Transitions collected while aiming at a particular goal can be used to learn about any other. With finite sets of goals, each transition is generally used to update the policy on every other goal \citep{mankowitz2018unicorn, kaelbling1993learning}. With continuous sets of goals, \emph{imaginary goals} are sampled from the goal space \citep{andrychowicz2017hindsight}. In the case of \uvfa policies, this consists in the substitution of the goal that is part of the input by the imaginary one, a technique called \emph{goal replay} or \emph{goal substitution}. Building on \uvfa, \citet{andrychowicz2017hindsight} proposed Hindsight Experience Replay (\her), a method leveraging hindsight for transferring knowledge between goals. The original goal of a transition can be substituted by any outcome experienced later in the trajectory (imaginary goal). This helps to increase the probability to observe rewards in reward-sparse environments. 
    
    In the literature, environments usually provide goals that the agent is asked to solve. In the \imgep framework however, autonomous agents are intrinsically motivated to set their own, possibly learning their representation \cite{laversanne2018curiosity}. \citet{forestier2016modular} in particular, biased the selection of the next goal to attempt towards modules showing high absolute measures of learning progress (LP). This mechanism helps the agent to focus on learnable goals and to disengage from goals that are currently too hard or already solved. \citet{veeriah2018many} uses LP computed from Bellman errors for goal selection, but this form of LP does not improve over random goal selection.
    
    Additional background can be found in the supplementary document. Table~\ref{sample-table} presents a classification of the multi-goal approaches most related to our work.
    
            
    \paragraph{Contributions.}The contributions of this paper are: 
    \begin{enumerate}
        \item A modular encoding of goals to enable learning of continuous sets of diverse goals within a single policy using \uvfa (Reach, Push, Pick and Place, Stack). This enables to tackle different kinds of goals, each with their own continuous parameterization, and facilitates transfer between modules and goals. See Sec. \ref{sec:multitaskgoal}.
        \item An active strategy for cross-module goal replay. Off-policy learning enables to use any experience to learn about any goal from any module. We propose to guide the selection of module for replay using absolute learning progress measures (in addition to LP-based goal sampling to interact with  environment). See Sec. \ref{sec:active_learning}.
        \item From the \imgep perspective, a single monolithic modular multi-goal action policy. This is an alternative to the population-based algorithms studied so far \cite{forestier2016modular, forestier2017intrinsically} and provides the flexibility of RL methods.
        \item An environment for modular goal RL. See Sec. \ref{sec:env}.
        \item Empirical comparisons to other architectures: a goal-parameterized RL with \her (flat multi-goal RL) and a multi-goal module-experts approach (one multi-goal expert per module). See Sec. \ref{sec:comparisons}.
        \item A study of the self-organization of learning phases demonstrated by our algorithm (automatic curriculum). See Sec. \ref{sec:details}.
        \item Experimental evidence of robustness to \emph{distracting goals}, \emph{forgetting} and \emph{body changes} in comparison with random goal selection. See Sec. \ref{sec:perturbations} and \ref{sec:distractors}.

    \end{enumerate}

\section{CURIOUS}

\subsection{A Modular Multi-Goal Architecture using Universal Approximators}
    \label{sec:multitaskgoal}
\begin{figure}[ht]
    \centering
        \includegraphics[width=0.8\columnwidth]{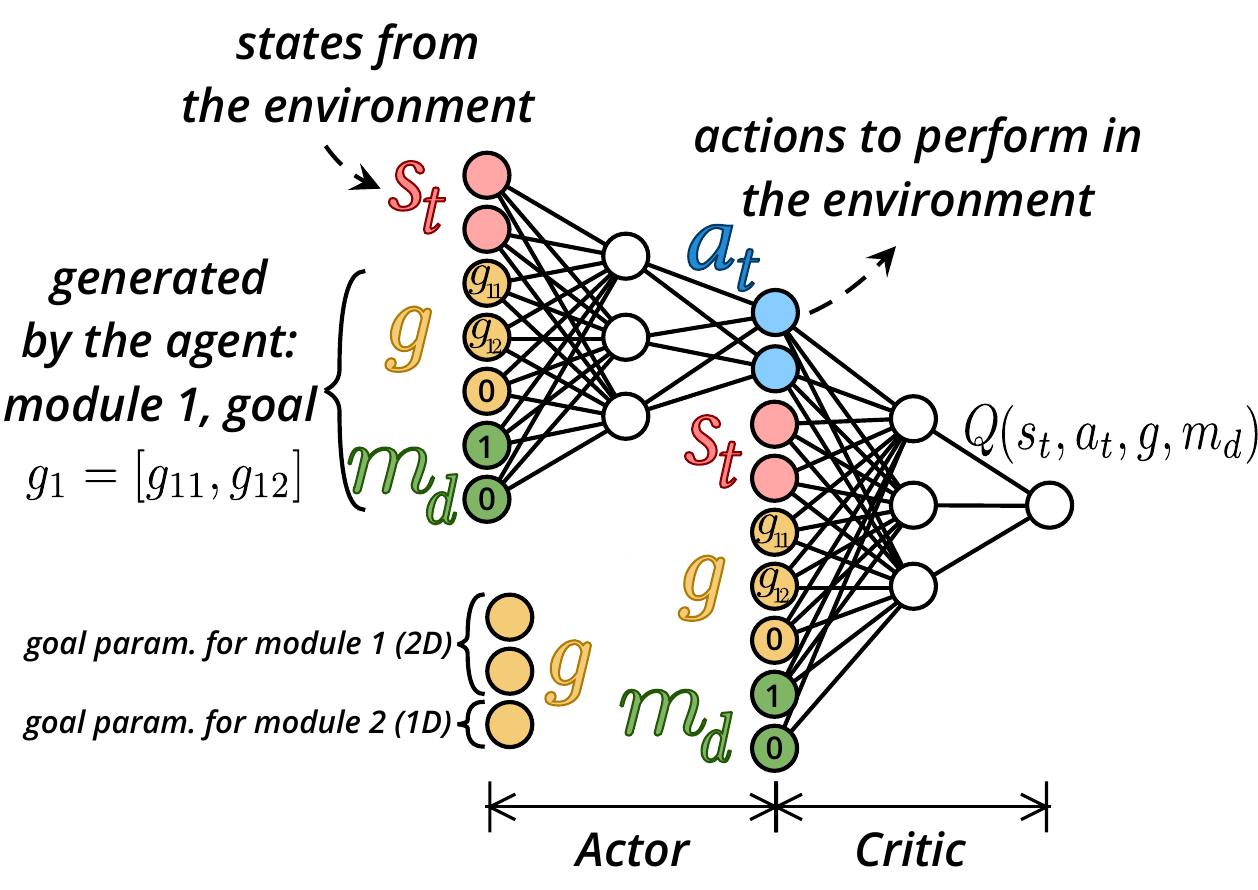}
         \caption{\textbf{Modular goal-parameterized actor-critic architecture (\muvfa)}. Toy example with $2$ modules, parameterized by $g_{1}$ (2D) and $g_{2}$ (1D) respectively. Here, the agent is attempting goal $g_1$ in module $M_1$, as specified by the one-hot module descriptor $m_d~=~\langle1,0\rangle$. The actor (left) computes the action $a_t$. The critic (right) computes the $Q$-value.}
    \label{fig:policy}
    \end{figure}

    \uvfa concatenates the goal of the agent with its current state to form the input of the policy and the value function implemented by deep neural networks \citep{schaul2015universal}. With \curious, we propose a new encoding of goals using modular representations. This enables to target a rich diversity of modular goals within a single network (\emph{modular multi-goal} approach), see Fig.~\ref{fig:policy}. Given $\mathcal{G}_{M_i}$ the goal space of module $M_i$, the current goal $g$ is defined as a vector of dimension $|\mathcal{G}| = \sum_{i=1}^N |\mathcal{G}_{M_i}|$, where the $\mathcal{G}_{M_i}$ can have different dimensionalities. $g$ is set to $0$ everywhere except in the indices corresponding to the current module $M_i$, where it is set to $g_i\in\mathcal{G}_{M_i}$. By masking the goal-inputs corresponding to unconsidered modules, the corresponding weights are frozen during backpropagation. In addition, a module descriptor $m_d$ of size $N$ (one-hot encoding) encodes the current module. The overall input to the policy network is $[s_t, g, m_d]$, see Fig.~\ref{fig:policy}. We call this modular goal-parameterized architecture Modular-\uvfa (\muvfa). 
    
    In Fig.~\ref{fig:policy}, we can see the underlying learning architecture (actor-critic). The actor implements the action policy and maps the input $[s_t,g,m_d]$ to the next action $a_t$. The action is then concatenated to a copy of the actor's input to feed the critic $[s_t, g, m_d, a_t]$. The critic provides an estimate of the $Q$-value: $Q(s_t,g,m_d,a_t)$. The critic and the actor are then trained using \ddpg \citep{lillicrap2015continuous}, although any other off-policy learning method could be used (e.g. \tdthree \citep{fujimoto2018addressing}, or \textsc{dqn} for the discrete case \cite{mnih2013playing}). More details about \ddpg can be found in the supplementary document or in \citet{lillicrap2015continuous}.

    \subsection{Module and Goal Selection, Cross-Module Learning, Cross-Goal Learning}
    \label{sec:active_learning}
    
    In \uvfa, \her and \unicorn, the next goal to target is selected at random \cite{schaul2015universal, andrychowicz2017hindsight, mankowitz2018unicorn}. This is coherent with the common view that the agent must comply with the desires of an engineer and target the goal it is asked to target. Here on the other hand, agents have the capacity to select which goal to target next. Because goals are not equivalent, the agent can benefit from intrinsic motivations towards LP \citep{schmidhuber1991curious, kaplan2004maximizing}. This can be useful: 1) when there are distracting goals on which the agent cannot progress; 2) when some goals are already mastered. This idea comes from the \imgep framework and was used in \citet{baranes2013active} to guide goal selection and in \citet{forestier2016modular} to guide module selection. The problem of selecting a module can be modeled as a non-stationary multi-armed bandit (MAB), where the value of each arm (module) is the current absolute LP. Learning progress (LP) is defined as the derivative of the agent's competence on a particular module: $LP_{M_i} = \frac{dC_{M_i}}{dt}$, where the competence $C_{M_i}:t\to p_{success}(t)$ is the probability of success at time $t$. Here, the agent focuses its attention on modules for which it is making the largest absolute progress, and pays little attention to modules that are already solved or unsolvable, i.e. for which $|LP|$ stays small. Using the absolute value of LP also leads to the prioritization of modules for which the agent is showing decreasing performances. This helps to deal with forgetting: the agent reallocates learning resources to the modules being forgotten, Sec. \ref{sec:perturbations}. 
    
     \paragraph{Learning Progress Estimation.} Since an autonomous agent is not externally provided its true competence or LP, it needs to approximate them for each module. To measure its competence, it uses some episodes (with $p_{eval}=0.1$) to evaluate itself on random modules and targets without exploration noise. The results (success $1$ or failure $0$) of these rollouts are stored in competence queues $results^{(i)}$ for all $M_i$. In a similar way as \citet{forestier2016modular}, the agent computes its subjective competence as 
    $$C_{M_i}(n^{(i)}_{eval}) = \frac{1}{l} \sum_{j=0}^{l-1} results^{(i)}(n^{(i)}_{eval}-j),$$
    where $n^{(i)}_{eval}$ is the number of self-evaluation rollouts performed by the agent in module $M_i$. The subjective $LP_{M_i}$ after $n^{(i)}_{eval}$ self-evaluation rollouts is then computed as:
    $$LP_{M_i}(n^{(i)}_{eval}) = C_{M_i}(n^{(i)}_{eval}) - C_{M_i}(n^{(i)}_{eval} - l).$$
    Given the subjective LP measures, we tackle the multi-armed bandit problem by implementing a simple approach called proportional probability matching, with an additional $\epsilon$-greedy strategy for exploration. More precisely, we compute the \emph{LP probabilities} $p_{LP}(M_i)$ as:
    $$p_{LP}(M_i) = \epsilon \times \frac{1}{N} + (1-\epsilon) \times \frac{|LP_{M_i}|}{\sum_{j=1}^{N} |LP_{M_j}|}, $$
    where $N$ is the number of modules. The ratio $\epsilon$ implements a mixture between random exploration of modules (left term) and exploitation through a biased selection/replay of modules (right term). The random exploration term enables sampling modules that do not show any LP (i.e. already solved, too hard, or at a plateau). This way, the agent can check that it stays competent on modules that are already learned, or can insist on modules that are currently too hard. 
    
    Note that we use LP for two distinct purposes: 1) Before data collection, to select the module from which to draw the next goal to attempt in the environment; 2) Before training, to select the substitute module descriptor (module replay). Recall that, once transitions are sampled from the replay buffer, they can be modified (replayed) by substituting the original module descriptor (or goal) by a new one. The substitute module is the one the agent is going to learn about. When replaying a particular module more than others, the agent allocates more resources to that module. While the use of LP for module selection is not new \cite{forestier2016modular}, we are the first to consider its use for cross-module goal replay. 
    
    \paragraph{Module and Goal Selection.} Before interacting with the environment, the agents selects the next goal to target by first sampling a module from $\mathcal{M}$ using $p_{LP}$, and second, sampling the goal uniformly from the corresponding goal space $\mathcal{G}_{M_i}$.
    
    \paragraph{Cross-Module and Cross-Goal Learning.} In an example with three modules, an agent computed $p_{LP} = [0.6, 0.2, 0.2]$. The agent uses these probabilities to guide learning towards modules with high absolute LP. If the size of the minibatch is $N_{mb}$, the agent will sample $\left \lfloor{N_{mb}\times 0.6}\right \rfloor$ transitions relevant to module $1$, $\left \lfloor{N_{mb}\times 0.2}\right \rfloor$ transitions relevant to module 2 etc. A transition that is \emph{relevant for module $M_i$} (e.g. Push module), means that it comes from an episode during which the corresponding outcome has changed (e.g. cube position). This sampling bias towards ``eventful'' transitions is similar to Energy-Based Prioritization \cite{zhao2018energy} (see supp. doc.). In this minibatch, every transition has been sampled to train on a specific module (e.g. $m_d^*$), although it could have been collected while targeting another module (e.g. $m_d$). To perform this cross-module learning, we simply substitute the latter by the former. Goal substitution is then performed using hindsight, which means the goal $g$ of a transition is sometimes ($p=0.8$) replaced by an outcome reached later in the same episode $g^*$ \cite{andrychowicz2017hindsight}. 
    
    \paragraph{Internal Reward.} After module descriptors and goals have been substituted, the agent computes an internal reward for each transition using a reward function parameterized by the new $m_d^*$ and goal $g^*$. Thus it answers: \emph{What would have been my reward for experiencing this transition, if I were aiming at that imagined goal from that imagined module?} The reward is non-negative ($0$) when the  outcome satisfies the constraints described by the imagined module $m_d^*$, relative to the imagined $g^*$; negative otherwise (-$1$). In a reaching module for instance (see Fig.~\ref{fig:mod_fetch}), a positive reward is generated when the Euclidean distance between the 3D target (goal) and the gripper (outcome) falls below a precision parameter $\epsilon_{reach}$ (reward constraint associated to the reaching module). 
    \subsection{Combining Modular-UVFA and Intrinsically Motivated Goal Exploration}

    A schematic view of \curious is given in Fig.~\ref{fig:schema_flow}. The detailed algorithm is given in the supplementary document.
    
    \begin{enumerate}
        \item \textbf{Module and goal selection.} The agent selects module $M_i$ and goal $g_i$ for the next rollout (blue), respectively sampled from the set of potential modules $\mathcal{M}$ using $p_{LP}$ (purple), and uniformly from the corresponding goal space $\mathcal{G}_{M_i}$.
        \item \textbf{Data collection.} The agent interacts with the environment using its current \muvfa policy (grey), collects transitions and stores them in memory (red).
        \item \textbf{LP update.} If it was a self-evaluation rollout, the agent updates its measures of competence, LP and $p_{LP}$ given the new result (success or failure, purple).
        \item \textbf{Module and goal substitution.} The agent decides on which modules and goals to train. To update the policy and critic, the algorithm first samples a minibatch from the replay buffers (red) using $p_{LP}$ and implements module and goal substitutions to perform cross-module and cross-goal learning (orange), see Sec.~\ref{sec:active_learning}.
        \item \textbf{Internal reward.} The agent computes its reward $r$ for each transition, using $R_{M,g}$ parameterized by the substitute module $m_d^*$ and goal $g^*$ (brown).
        \item \textbf{RL updates.} The agent updates its policy and value function with \ddpg using the modified minibatch (green). 
    \end{enumerate}
    
    \begin{figure}[ht]
      \centering  
      \includegraphics[width=0.8\columnwidth]{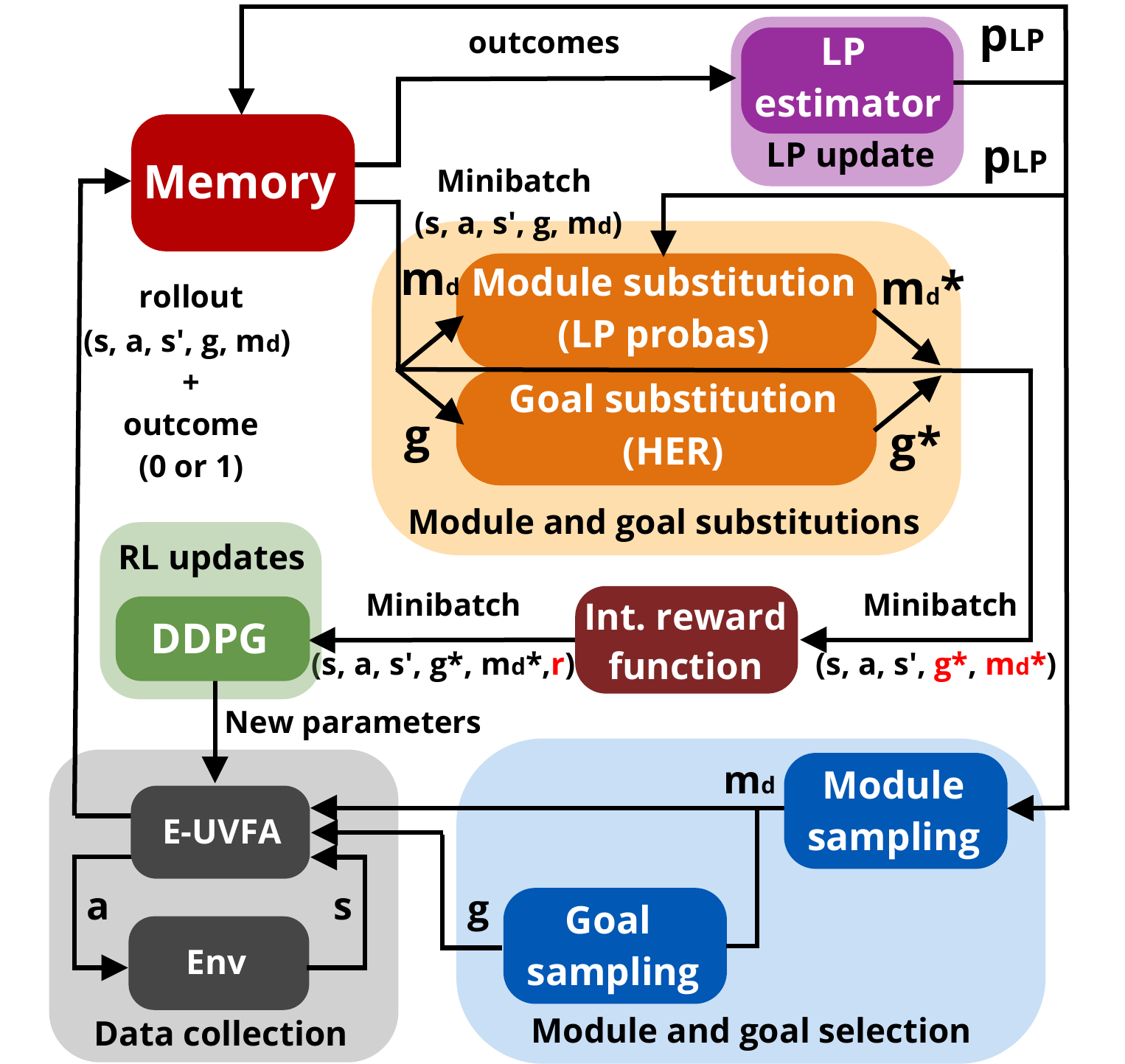}
      \caption{\textbf{Schematic view of \curious.}}
      \label{fig:schema_flow}
    \end{figure}

    \subsection{Evaluation Methodology}
    The performance of the agents are evaluated offline in terms of success rates over sets of achievable goals (defined by the experimenter). Every point of a learning curve represents the success rate over $95$ offline evaluation rollouts ($5\times 19$ actors), using random achievable goals. Evaluation is completely independent from training, i.e. agents cannot train on evaluation transitions. We use the non-parametric one-tail Mann-Whitney U-test with confidence level $\alpha~=~0.01$ for all comparisons. More details and justifications can be found in the supplementary document.

\section{A Modular Goal Environment}
    \label{sec:env}

    \emph{Modular Goal Fetch Arm} is a new simulated environment adapted from the OpenAI Gym suite \citep{brockman2016openai}. The agent is embodied by a robotic arm facing 2 cubes randomly positioned on a table.  The agent controls the position of its gripper and the gripper opening ($4$D). It can target a diverse set of modular goals: ($M_1$) \emph{Reach} a $3$D target with the gripper; ($M_2$) \emph{Push} cube $1$ onto a $2$D target on the table; ($M_3$) \emph{Pick and Place} cube $1$ on a $3$D target; ($M_4$) \emph{Stack} cube $1$ over cube $2$. Additional Push modules concerning additional out-of-reach and moving cubes can be defined (impossible, distracting goals). Further details can be found in the supplementary document.

\section{Experiment and Results}
In this section, we present ablative studies to assess the relative importance of: 1) the policy and value function architecture and 2) the use of intrinsically motivated module selection for practice and replay. We call \muvfa the algorithm using a modular goal-parameterized policy and random module choices, while the intrinsically motivated version is called \curious. We do not investigate the efficiency of \her or the efficiency of the sampling bias towards interesting transitions as they were already studied in \citet{andrychowicz2017hindsight, plappert2018multi} and \citet{zhao2018energy} respectively. For fair comparisons, we apply both mechanisms to all the tested algorithms.

    \begin{figure*}[!hb]
      \centering
      \subfigure[\label{fig:c}]{\includegraphics[width=0.323\textwidth]{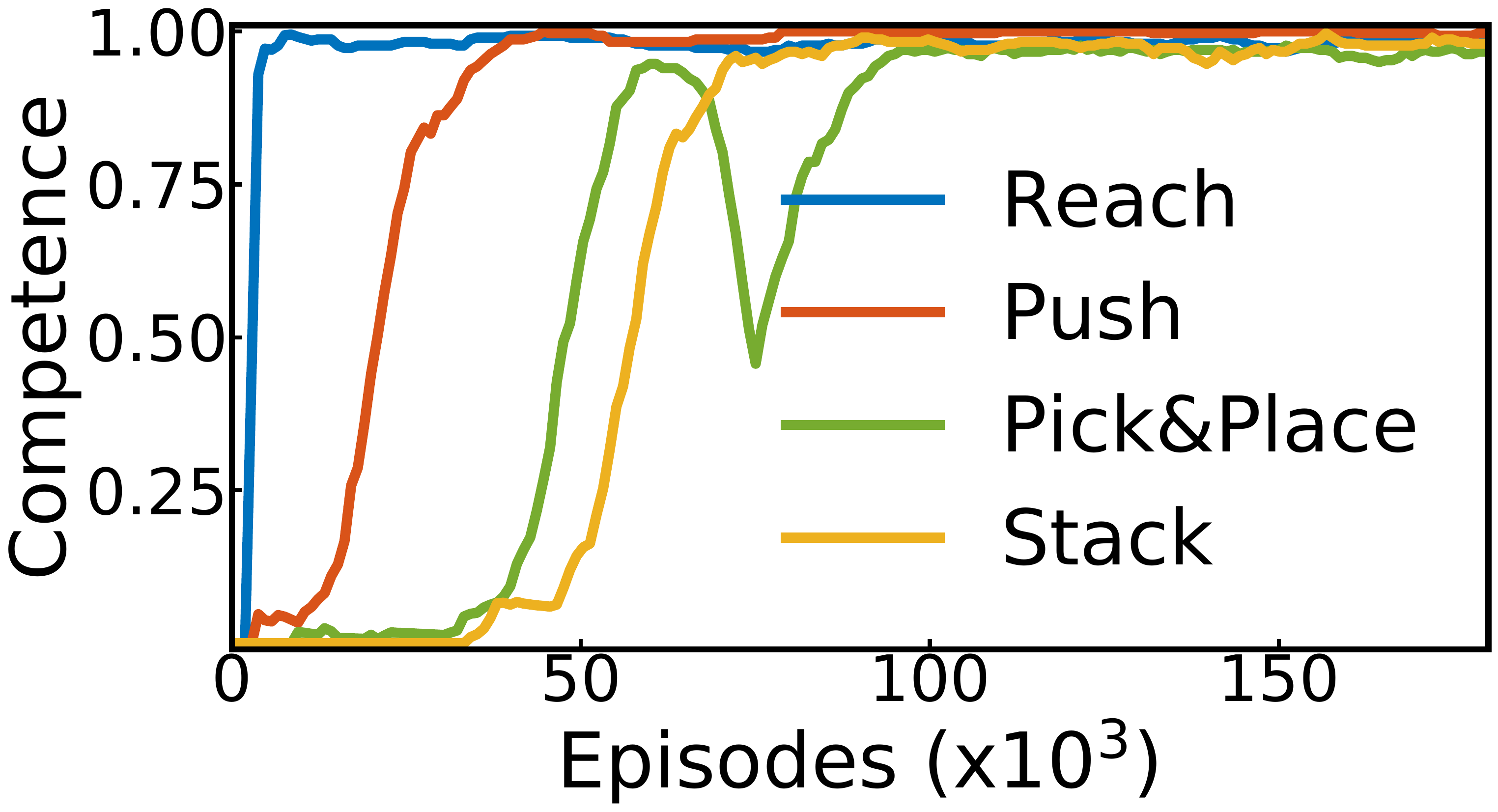}}
      \subfigure[\label{fig:cp}]{\includegraphics[width=0.32\textwidth]{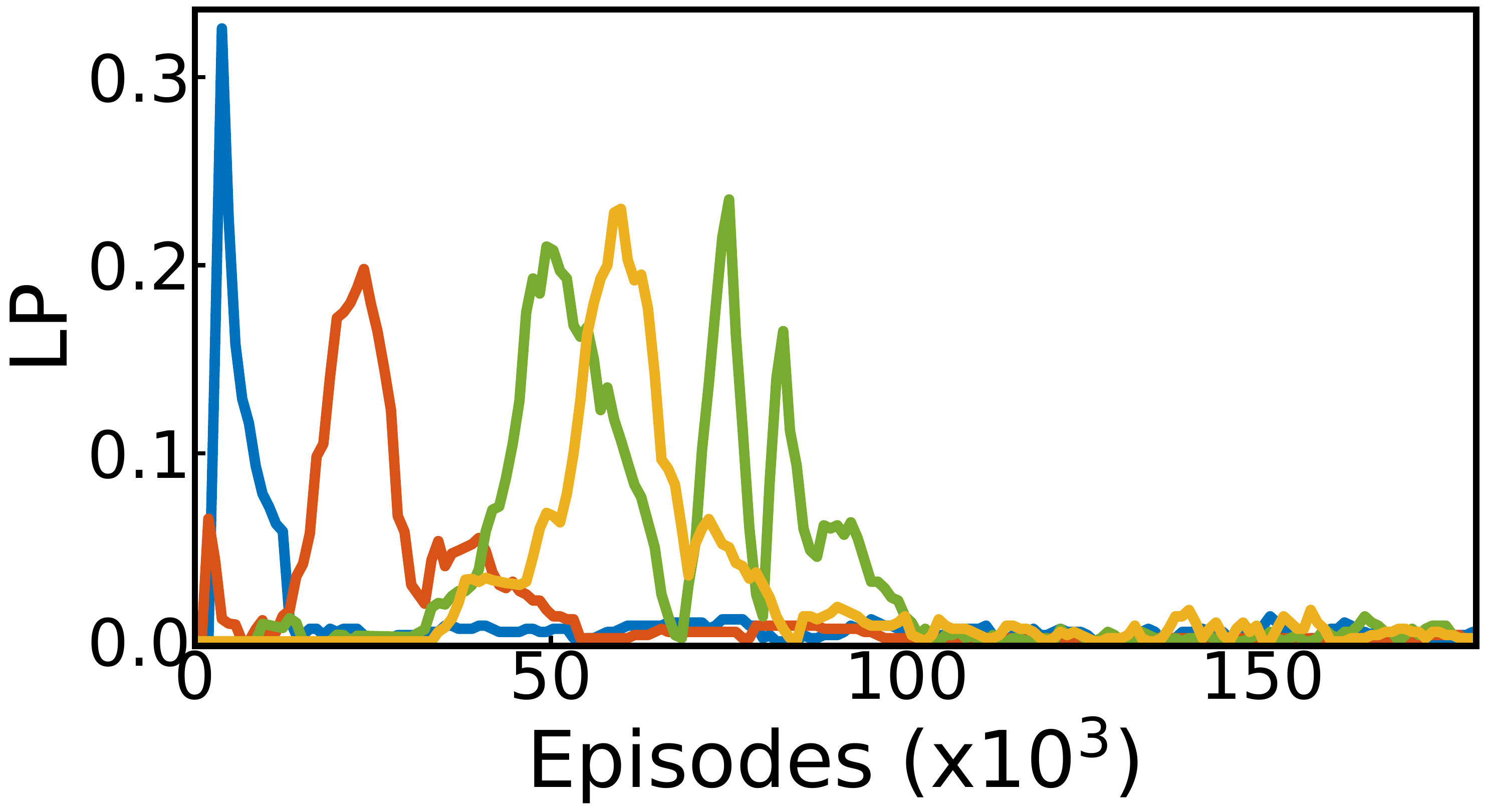}}
      \subfigure[\label{fig:proba}]{\includegraphics[width=0.323\textwidth]{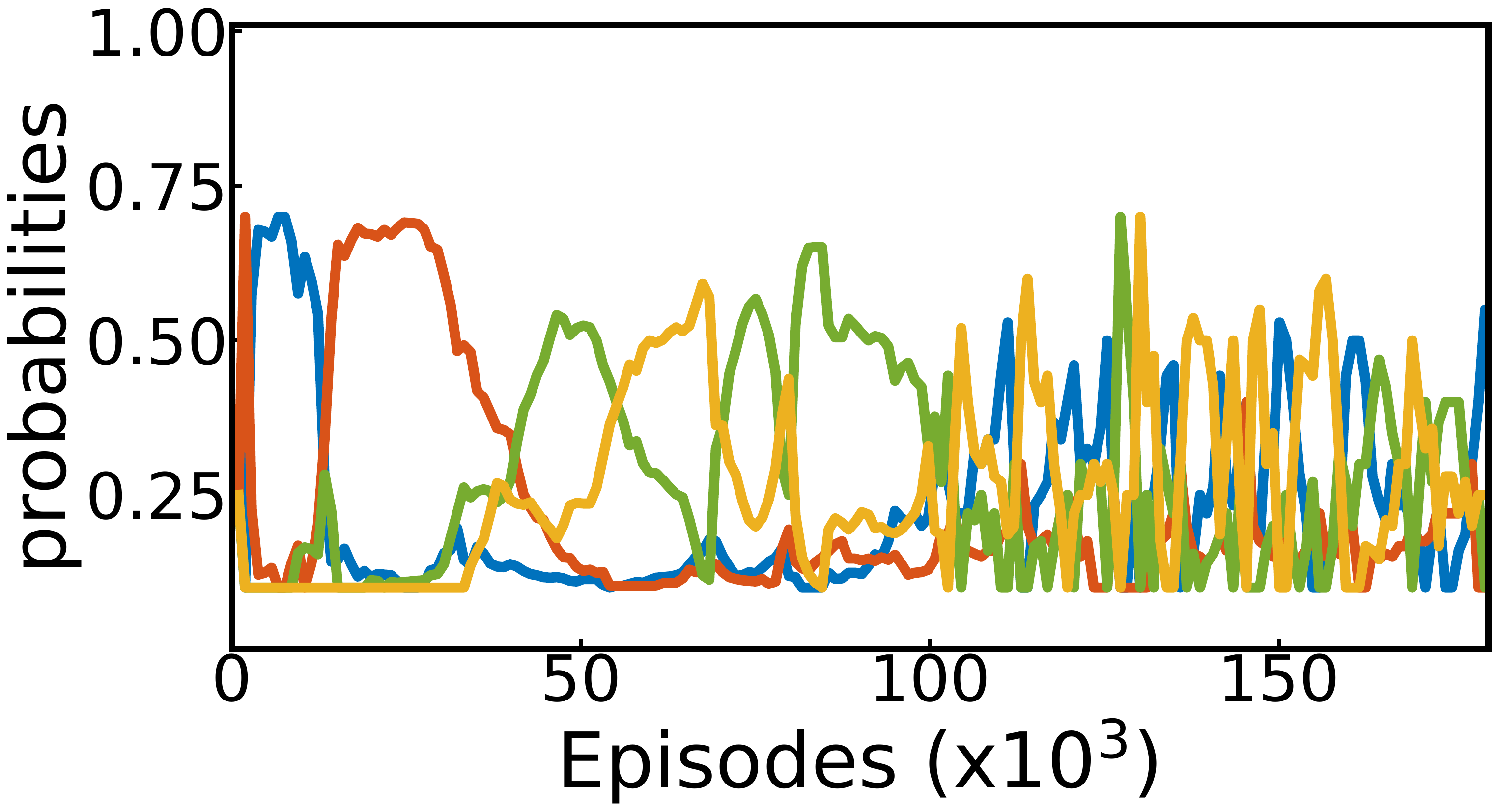}}
      \caption{\textbf{Visualization of a single run.} a: Module-dependent subjective measures of competence for \curious (1 run). b: Corresponding module-dependent subjective measures of absolute LP. c: Corresponding probabilities $p_{LP}$ to select modules to practice or to learn about. }
      \label{fig:figures}
    \end{figure*}   
    
    \begin{figure}[ht]
        \centering
        \includegraphics[width=\columnwidth]{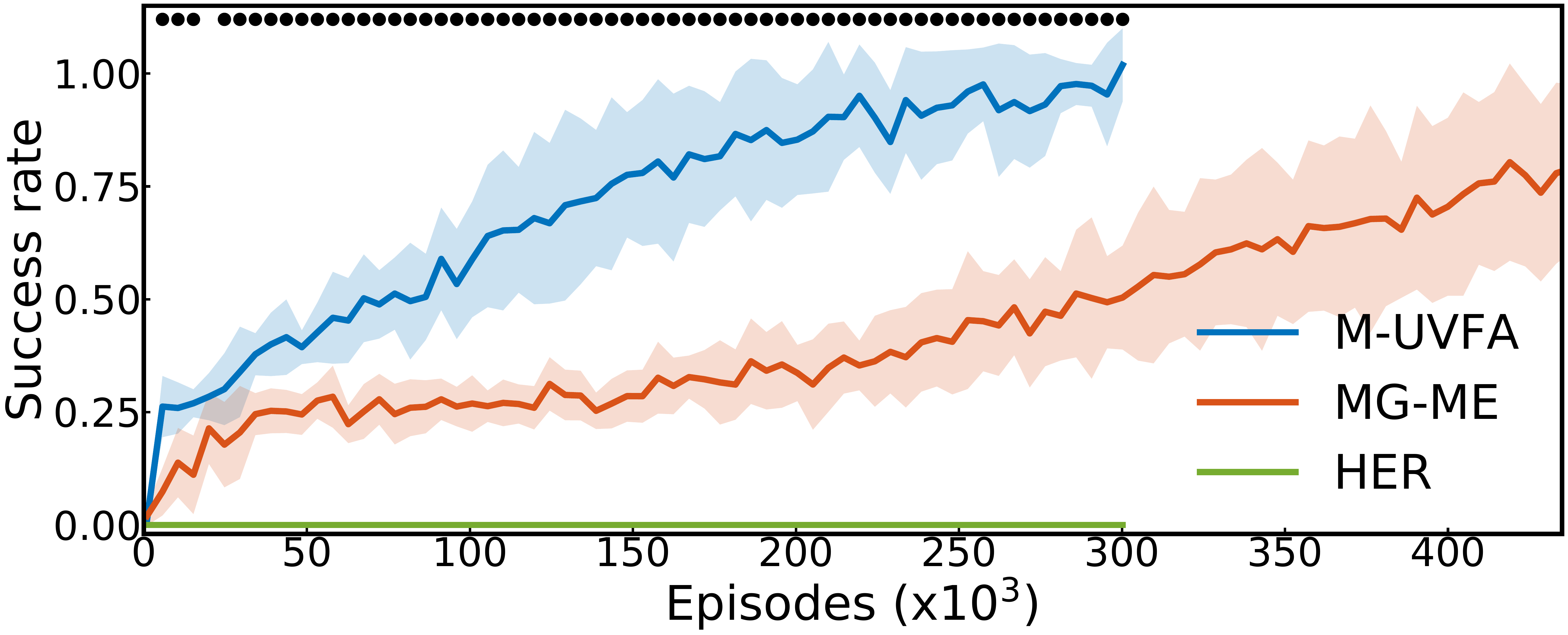}
        \caption{\textbf{Impact of the policy and value function architecture.} Average success rates computed over achievable goals. Mean +/- std over 10 trials are plotted, while dots indicate significance when testing \muvfa against \te.} 
        \label{fig:success_rates}
    \end{figure}
    
        
    \subsection{Impact of Policy and Value Function Architecture}
    \label{sec:comparisons}
    \paragraph{Experiments.} In this section, we investigate the impact of using an \muvfa architecture for the policy and value function. The module-set is composed of four achievable modules and four distracting modules. We test this algorithm against two baselines: 
    \begin{enumerate}
        \item A \emph{flat multi-goal} architecture (\her). This algorithm does not represent goals in a modular fashion but in a linear way. The corresponding goal is selected uniformly inside $\mathcal{G}$, a holistic goal space such that $\mathcal{G} = \prod_{i=1}^N \mathcal{G}_{M_i}$. To generate a reward, the agent needs to satisfy the constraints described by all the modules at once. This goal-parameterized architecture is equivalent to \uvfa, which makes the algorithm equivalent to \her+\ddpg.
        \item A \emph{multi-goal module-experts} architecture (\te) where an expert multi-goal policy is trained for each of the $N$ modules. Each policy is trained one epoch every $N$ on its designated module and shares the collected transitions with other experts. When evaluated on a particular module, the algorithm uses the corresponding module-expert. 
    \end{enumerate}

    \paragraph{Results.}
    Fig.~\ref{fig:success_rates} shows the evolution of the average success rate computed over achievable goals for \muvfa and the two baselines described above. The learning curve of \her stays flat. This can be easily understood as none of the goals expressed in the complete goal space $\mathcal{G}$ corresponds to a real situation (e.g. the agent cannot reach a 3D target with its gripper while placing a cube at another). The agent cannot fulfill the constraints of all modules simultaneously, thus receives no reward. This motivates the use of a modular representation with separated modules. Comparing \te and \muvfa, we can see that the achievable goals are learned much faster in the multi-modular-goals approach (one, policy, $\approx 250\cdot10^3$ vs. $\approx 450\cdot10^3$ episodes). From now on, all experiments use the \muvfa architecture.

    %


    \subsection{Visualizing the Intrinsic Motivation towards Learning Progress}
    \label{sec:details}
    \paragraph{Experiments.} This section aims at showing the inner working of \curious's intrinsic motivation towards LP. Here we focus on a setting with four achievable modules (Reach, Push, Pick and Place, and Stack). 
    
    \paragraph{Results.} Fig.~\ref{fig:c} shows the evolution of the module-dependent competence measures as subjectively perceived by the agent, while Fig.~\ref{fig:cp} shows the evolution of the corresponding LP measures. Finally, Fig.~\ref{fig:proba} shows the corresponding module selection probabilities $p_{LP}$, a mixture of random selection with probability $\epsilon$ and active selection proportional to LP measures with probability $1-\epsilon$. These figures demonstrate the existence of successive learning phases, that can be interpreted as developmental phases \cite{oudeyer2016evolution}. The robot first learns how to control its gripper ($M_1$), then to push objects on a desired target on the table ($M_2$) before it learns how to place the cube on a $3$D target ($M_3$) and how to stack the two cubes ($M_4$). Fig.~\ref{fig:cp} shows that LP stays small for modules that are already solved (e.g. $M_1$ after $10^4$ episodes) or too hard to solve (e.g. $M_3$ and $M_4$ before $35\cdot10^3$ episodes), and increases when a module is being learned. We further discuss the link between these learning phases, developmental learning and curriculum learning in the supplementary document.


  \subsection{Impact of the Intrinsic Motivation: Resilience to Forgetting and Sensor Perturbations}
    \label{sec:perturbations}
    \paragraph{Experiments.} During learning, the agent can forget about a previously mastered module. This can happen because is not targeting it often (catastrophic forgetting), because of environmental changes (e.g. icy floor) or because of body changes (e.g. sensor failure). Ideally, \curious should be able to detect and react when such situations arise. This section investigates the resilience of our algorithm to such perturbations and compares it to the \muvfa baseline.
    
    We first look at a run where forgetting occurs and explain how \curious detects the situation and reacts. Since forgetting cannot be triggered, we add more emphasis to a second experiment, where we simulate a time-locked sensory failure. We present the following setup to the agent: first, it learns about a set of $4$ modules (Reach, Push, Pick and Place for cube $1$, and Push for cube $2$). Then, a sensory perturbation is triggered at a precise time ($epoch=250$, $episode=237.5\cdot10^3$) such that the perception of cube $2$ gets shifted by $0.05$ (simulation units) until the end of the run. The performance on this module suddenly drops and we compare the recoveries of \curious and \muvfa.

    \paragraph{Results - Forgetting.} Looking at Fig.~\ref{fig:c}, we can observe a drop in the competence on $M_3$ around episode $80\cdot10^3$. This phenomenon is usually described as \emph{catastrophic forgetting}: because it is trained on other modules, the network can forget about the previously mastered module $M_3$, without any obvious reason. The corresponding period of Fig.~\ref{fig:cp} shows an increase in LP for $M_3$, which in turn triggers an additional focus of the agent towards that module (see the corresponding probability increase in Fig.~\ref{fig:proba}). Using LP to bias its attention, the agent monitors its competence on the modules and can react when it forgets about a previously mastered module. This mechanism helps to deal with the problem of forgetting and facilitates learning of multiple modules in parallel. To prove its efficiency, we need to compare \curious to its baseline \muvfa using a time-locked perturbation.
    
    \begin{figure}[ht]
        \centering
        \includegraphics[width=0.9\columnwidth]{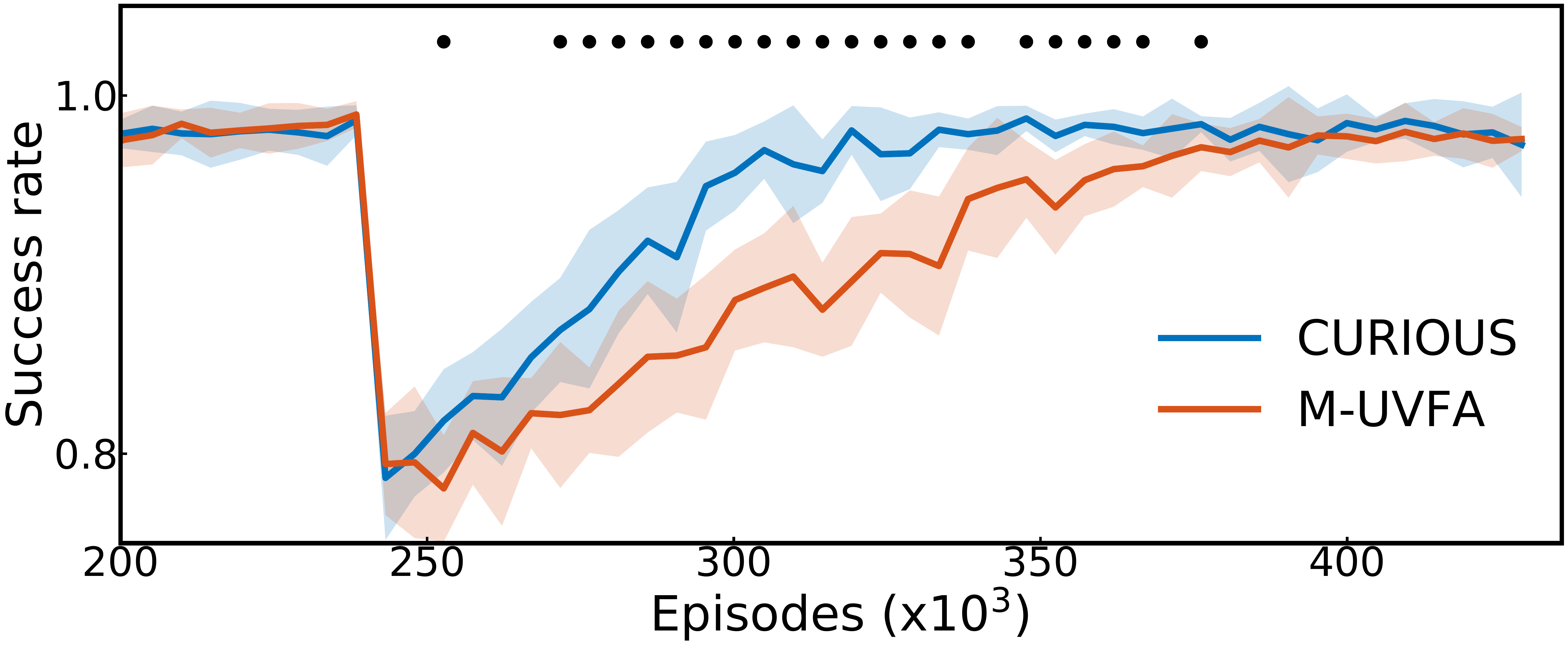}
        \caption{\textbf{Impact of the intrinsic motivation towards LP for sensory failure recovery.} Mean success rates over the four modules +/- std over 10 trials are plotted. The dashed line indicates the onset of the perturbation, while the dots indicate significance when testing \curious against \muvfa.}
        \label{fig:perturb}
    \end{figure}
    
    \paragraph{Results - Sensor Perturbation.}In Fig.~\ref{fig:perturb}, we can observe the drop in average success rate after the perturbation (around $240\cdot10^3$ episodes). This perturbation only affects one of the four modules (Push cube 2), which triggers a drop in the average performance of about $1/4^{th}$. As described above, \curious agents detect that perturbation and react by using more transitions to improve on the corresponding module. This translates into a significantly faster recovery when compared to \muvfa. The agents recover $95\%$ of their pre-perturbation performance in $78$ and $43 \cdot 10^3$ episodes (random and active respectively), which translates in a $45\%$ faster recovery for \curious ($p<10^{-4}$), see Fig.~\ref{fig:perturb}.


    \subsection{Impact of the Intrinsic Motivation: Resilience to Distracting Modules}
    \label{sec:distractors}
    
    \paragraph{Experiments.}In this section, we investigate the resilience of our learning algorithm when the number of distracting modules increases (0, 4, 7). The agent faces four achievable modules in addition to the distracting modules. The distracting modules are all Push modules relative to the randomly moving and out-of-reach cubes. The agent receives extra noisy inputs corresponding to the random movements of these cubes.

    \begin{figure}[ht]
        \centering
        \includegraphics[width=0.9\columnwidth]{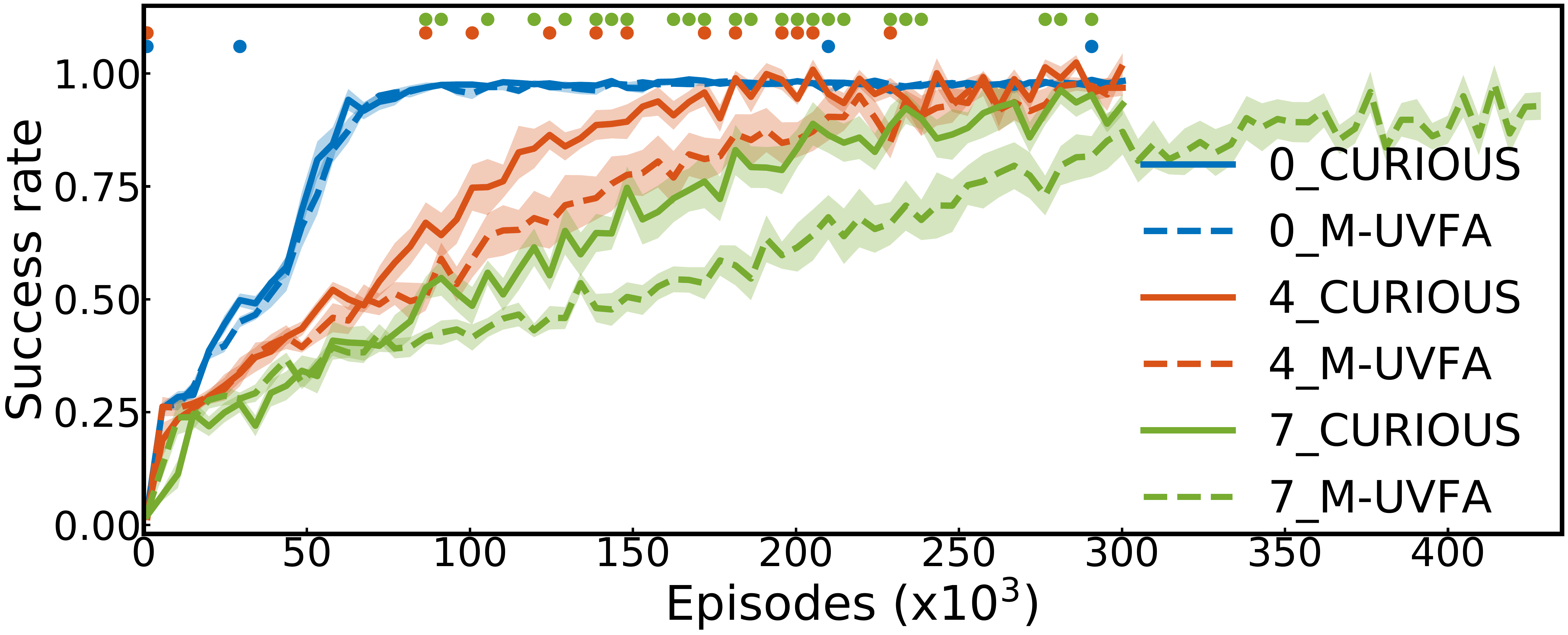}
        \caption{\textbf{Impact of the intrinsic motivation towards LP when the number of distracting modules grows.} Average success rates computed over achievable goals when the number of distracting modules increases (0, 4, 7). Mean and standard error of the mean (for visualization purpose) over 10 trials are plotted. The dots indicate significance when testing \curious against \muvfa. }
        \label{fig:distractors}
    \end{figure}

    \paragraph{Results.} In Fig.~\ref{fig:distractors}, we see that the number of distracting modules faced by the agents highly impacts their learning speed on achievable goals. In particular, \muvfa random agents do not know that these goals are impossible to achieve and waste time and resources trying to improve on them. Since these agents sample distracting modules just like others, we can expect the learning speed to be scaled by $\frac{\# achievable modules}{\#modules}$. On the other hand, \curious agents try to learn which modules are too difficult at the moment to target them less often. Note that \curious agents still need to choose them sporadically to keep updated measures of their LP: they sample a random module with probability $\epsilon$. In Fig.~\ref{fig:distractors}, we see that the advantage of \curious over its random counterpart increases as the number of distracting modules grows (see colored dots indicating significant differences). Although the addition of distracting modules might sound a bit ad-hoc here, it is important to note that autonomous agents evolving in the real world face numerous modules such as these. For humans, quantity of potential modules are impossible (predicting the movement of leaves on a tree, trying to walk on walls etc.). Just as humans, artificial agents need to discard them based on experience and LP.


\section{Discussion}
    \label{sec:discu}

    \textbf{Leveraging Environment Modularity.} In some environments, representing all the potential goals requires modular representations. Because Sec.~\ref{sec:comparisons} proved that a simple \uvfa architecture could not deal with this situation, we proposed \muvfa. Note that, although our Modular Goal Fetch-Arm environment only contains goals that can be represented in a modular way, \muvfa can also target discrete sets of goals using flat representations (by setting the goal $g_i$ of module $M_i$ to the null vector and letting $m_d$ encode for the goal). In short, \muvfa enables traditional \uvfa to target a richer diversity of goals than what was possible with traditional \uvfa implementations.
    
    \paragraph{Pros and Cons of Monolithic Policies.}As noted in \citet{mankowitz2018unicorn}, representations of the world state are learned in the first layers of a neural network policy/value function. A representation learned to achieve goals from one module could probably be useful for learning goals from another similar module. Our monolithic modular goal policy leverages that fact, by re-using subparts of the same network to learn different but similar modules and goals. This might explain why \muvfa outperforms the \emph{multi-goal module-experts} (\te) policy architecture (Fig.~\ref{fig:success_rates}). However, such monolithic policies are more prone to forgetting. Although this phenomenon is partially mitigated by the use of the absolute value of LP, it might still be an issue when the number of potential modules increases. To answer this problem, we could think of combining several \muvfa policies for different subsets of modules.
    
    \textbf{A Monolithic IMGEP.} Contrary to the vision shared by many multi-goal RL papers where agents must comply to the engineer desires (\emph{do goal 1, do goal 3 ...}), our work takes the perspective of agents empowered by intrinsic motivations to choose their own goals (\emph{do whatever you want, but be curious.}). This vision comes from the \imgep framework which defines agents able to set their own parameterized problems to explore their surrounding and master their environment \citep{forestier2017intrinsically}. Contrary to previous \imgep algorithms grounded on memory-based representations of policies, \curious uses a single monolithic policy for all modules and goals (\muvfa). Because it is memory-based, \textsc{macob} does not handle well the variety of initial states which limits its generalization capacity. 
    
    \paragraph{Active Learning using Learning Progress.}Although LP-based module selection already brings significant advantages compared to random module selection, \curious could benefit from a more advanced LP estimator. Our current estimator uses moving averages. It is fast and requires small amounts of memory, but could be more reactive to changes in true LP. This delay causes the agent to persevere on modules that are already mastered, or not to react quickly to newly learnable modules. These drawbacks could be mitigated with more advanced measures of competence or LP (e.g. approximate Bayesian methods like in \citet{mathys2011bayesian}).
    
    
    \section{Further Work}

    \textbf{Hierarchical Extension.} The idea of using a high-level policy to select goals for a lower-level policy was also studied in the field of \emph{hierarchical RL}. Yet, while hierarchical RL agents choose their own subgoals, they usually do so to achieve higher-level goals imposed by the engineer \cite{vezhnevets2017feudal,nachum2018data,levy2018hierarchical}. A natural extension of our work could be to replace our high-level MAB module selection policy by another \curious agent targeting self-generated higher-level goals, in a hierarchical manner.
        
    \paragraph{Learning a Goal Selection Policy.} In this work we provide the policy for goal sampling inside modules: sampling uniformly from a pre-defined (reachable) goal space. In the future, the agents could learn it autonomously using adaptations of existing algorithms such as \saggriac \cite{baranes2013active} or \textsc{goal-gan} \cite{held2017automatic}. \saggriac enables to split recursively a wide continuous goal space and to focus on sub-regions where LP is higher, while \textsc{goal-gan} proposes to generate goals of intermediate difficulty using a Generative Adversarial Network.
    
    
    \paragraph{Learning Representations for Modules and Goals.} Another assumption of our work, is that agents should already know a modular representation of goals and their modules. Although modules and goal spaces were hand-defined in the experiments of this paper, this was a scaffolding for the studies we presented. In a general IMGEP setting,  autonomous agents must be able to construct their own set of modules and goal representations. The idea of autonomously learning modular goal representations from experience has been explored in \citet{laversanne2018curiosity}, using $\beta-VAEs$.  This was used for goal exploration using a population-based IMGEP algorithm. Combining \curious to this unsupervised learning of disentangled goal spaces is an interesting avenue to explore, in the quest of more autonomous learning agents. 
    
    \section{Conclusion}

    This paper presents \curious, a learning algorithm that combines an extension of \uvfa to enable modular goal RL in a single policy (\muvfa), and active mechanisms that bias the agent's attention towards modules where the absolute LP is maximized. This self-organizes distinct learning phases, some of which are shared across agents, others dependent on the agent experience. With this mechanism, agents spend less time on impossible modules and focus on achievable ones. It also helps to deal with forgetting, by refocusing learning on modules that are being forgotten because of model faults, changes in the environment or body changes (e.g. sensory failures). This mechanism is important for autonomous continual learning in the real world, where agents must set their own goals and might face goals with diverse levels of difficulty, some of which might be required to solve others later on.

    \paragraph{Links.}The \href{https://github.com/flowersteam/gym_flowers}{environment}, \href{https://github.com/flowersteam/curious}{code} and \href{https://www.youtube.com/watch?v=qO_OZpsXXGQ&feature=youtu.be}{video} of the \curious agent are made available at \url{https://github.com/flowersteam/curious}.

 \subsection*{Acknowledgments}

 C\'edric Colas is partly funded by the French Minist\`ere des Arm\'ees - Direction G\'en\'erale de l’Armement. Olivier Sigaud is partly funded by the European Commission, within the DREAM project. The DREAM project has received funding from the European Unions Horizon 2020 research and innovation program under grant agreement $N^o$ 640891. 

\bibliography{biblio}
\bibliographystyle{icml2019}

\section*{Additional Background}

        \paragraph{DDPG.} Deep Deterministic Policy Gradient (DDPG) is an off-policy model-free RL algorithm for continuous action spaces \cite{lillicrap2015continuous}. It concurrently learns a policy $\pi$ (the actor) and a $Q$-function $Q$ (the critic) using neural networks. The actor implements the controller and maps the current state to the next action: $\pi: \mathcal{S} \to \mathcal{A}$. The critic approximates the optimal action-value function $Q^*$, $Q:\mathcal{S}\times\mathcal{A} \to \mathbb{R}$. 
        
        For exploration purposes, transitions are collected using a noisy version of the policy (behavioral policy), e.g. using Gaussian noise on the actions. The transitions are stored in a replay buffer of finite size. The actor and critic updates are then conducted using transitions sampled uniformly from the replay buffer. The critic is trained to minimize the mean-squared Bellman error such that:
        
        $$L(\phi, \mathcal{D}) = E_{t\sim\mathcal{D}} \bigg[\Big(Q_\phi(s,a) - \big(r + \gamma \max\limits_{a'} Q_\phi(s', a')\big)\Big)^2 \bigg],$$
        
        where $\phi$ represents the parameters of the critic, $\mathcal{D}$ is the dataset, $t=(s,a,r,s')$ is a transition with $r$ a reward. The actor is trained so as to maximize the output of the critic. Further details can be found in the original paper \cite{lillicrap2015continuous}.
        
        \paragraph{UVFA.} Universal Value Function Approximators (\uvfa) is an extension of the traditional actor and critic networks to target multiple goals \cite{schaul2015universal}. The goal is represented by a vector defined in space $\mathcal{G}$. It is concatenated both to the input of the actor and to the input of the critic such that: $\pi:\mathcal{S} \times \mathcal{G} \to \mathcal{A}$ for the actor and $Q:\mathcal{S} \times \mathcal{G} \times \mathcal{A} \to \mathbb{R}$ for the critic. The reward function as well, is now parameterized by the goal: $R_g:\mathcal{S} \times \mathcal{A} \to \mathbb{R}$. The original paper shows that a \uvfa architecture trained with Deep Q-Network (\textsc{dqn}) can generalize to previously unseen goals \cite{schaul2015universal}. Further details can be found in the original paper.
        
        \paragraph{HER.} Hindsight Experience Replay (\her) proposes an efficient mechanism leveraging hindsight to learn more efficiently in multi-goal settings \cite{andrychowicz2017hindsight}. When an agent makes an attempt towards a goal and fails to reach it, \her proposes to probabilistically substitute the goal vector by an outcome achieved later in the trajectory. In simpler words, the agents \emph{pretends} it was trying to reach some other goal it actually reached later on, hence it increases the probability to observe a reward. Note that this mechanism requires the agent to have access to the true reward function $R_g$ and to be able to compute the reward corresponding to any combination of goal, state and action vectors. This mechanism can be used with any off-policy algorithms and was initially presented with \textsc{dqn} and \textsc{ddpg}. Further details can be found in the original paper \cite{andrychowicz2017hindsight}.
        
        \begin{figure}[ht]
          \centering  
          \includegraphics[width=\columnwidth]{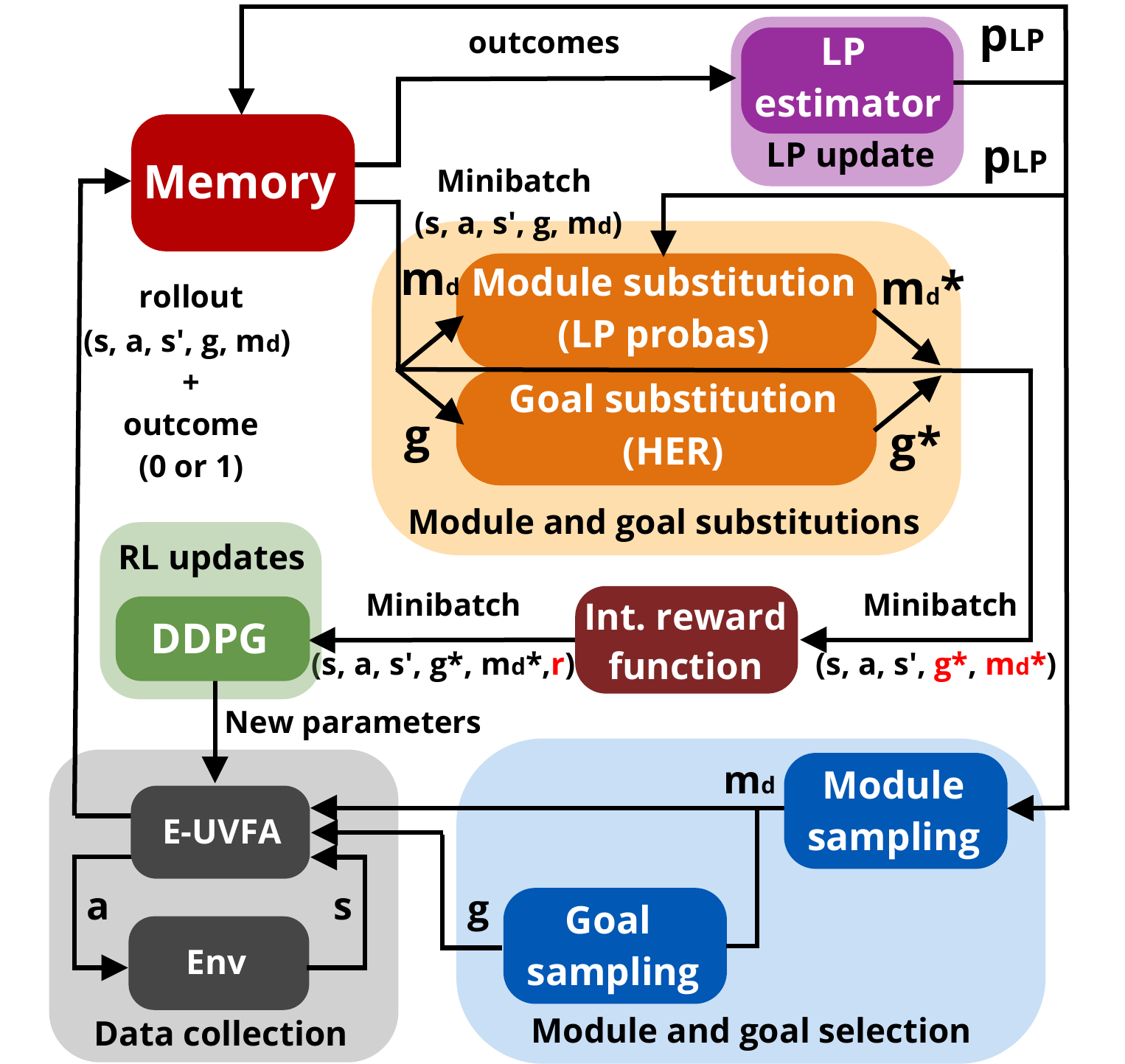}
          \caption{\textbf{Schematic view of \curious.}}
          \label{fig:schema_flow}
        \end{figure}

    \section*{Additional Methods}

            \begin{algorithm*}[!hb]
            \caption{CURIOUS}
            \label{alg:example}

            \begin{algorithmic}[1]
               \STATE \textbf{Input:} env, $\mathcal{M}$, $\mathcal{G}_{1:N}$, $noise$, internal\_reward(~), $p_{eval\_c}$ \hspace{1.cm} \COMMENT{$\mathcal{M}$: set of $N$ modules, $\mathcal{G}_i$: goal space of module $M_i$ } 
               \STATE \textbf{Initialize:} $policy$, $memory$, $p_{LP}$  \hspace{5.7cm} \COMMENT{random, empty and uniform respectively}
               \REPEAT 
                   \STATE $eval\_c \gets random() < p_{eval\_c}$ \hspace{5.3cm} \COMMENT{The agent evaluates its competence if True}
                   \STATE $goal, m_d(M_i) \gets $ \textbf{ModuleGoalSelector()} \hspace{2.8cm} \COMMENT{$M_i$ $\sim p_{LP}$ (or uniform if $eval\_c$), $goal$ $\sim   \mathcal{U}(\mathcal{G}_{M_i})$}
                   \STATE $s_0, outcome_0 \gets $ env.reset()
                    \FOR{$t=0:N_t$}
                        \STATE $policy\_input \gets$ concatenate($s_t, m_d, goal$)
                        \STATE $a_t \gets  policy(policy\_input)$
                        \IF{not $eval\_c$}
                            \STATE $a_t \gets a_t + noise$ \hspace{3cm}   
                        \ENDIF
                        \STATE $s_{t+1}, outcome_{t+1} \gets$ env.step($a_t$) 
                  \ENDFOR
                        \STATE $memory$.add($s, a, s', goal, outcome, m_d$)
                        \IF{$eval\_c$} 
                            \STATE $memory$.add($outcome==goal$, $m_d$)   \hspace{3.6cm} \COMMENT{If self evaluation, update LP with new result}
                            \STATE $p_{LP} \gets$ \textbf{LP\_Update(} $memory$ \textbf{)}
                        \ENDIF
                        \STATE $modified\_batch \gets$ \textbf{ModuleGoalReplayPolicy(}$p_{LP}$, $memory$\textbf{)} \hspace{0cm} \COMMENT{Use Prioritized Interest Replay, $p_{LP}$ and \her}
                        \STATE $modified\_batch \gets$ \textbf{RewardComputer(}$modified\_batch$, internal\_reward()\textbf{)}
                        \STATE $policy \gets$ \textbf{RL\_Updates(}$modified\_batch$\textbf{)} \hspace{7.8cm} \COMMENT{Using \ddpg}
               \UNTIL{learning over}
            \end{algorithmic}
           
        \end{algorithm*}

        \paragraph{Overview.} Algorithm 1 and Fig.~\ref{fig:schema_flow} present the pseudo-code. We go through the different steps:

        \begin{enumerate}
            \item \textbf{Module and goal selection.} The agent selects module $M_i$ and goal $g_i$ for the next rollout (blue in Fig.~\ref{fig:schema_flow}). First, it decides whether it will perform self-evaluation, using probability $p_{eval\_c}$ [line 4 in the algorithm]. If it does so, it selects a module $M_i$ at random. Otherwise, the next module is sampled from the set of modules $\mathcal{M}$ according to the LP probabilities $p_{LP}$ (purple). The goal is sampled uniformly from the corresponding goal space $\mathcal{G}_{M_i}$ [line 5]. All the considered goal spaces are either the 2D Euclidean space (for the \emph{Push module}, with $z=\space$~height(table)) or the 3D Euclidean space (for other modules). These spaces are simply subspaces of the state space $\mathcal{G}_{ \subset \mathcal{S}}$.
            
            \item \textbf{Data collection.} The agent interacts with the environment and collects transitions (grey) using its current \muvfa policy [lines 8 to 13] before storing them in memory [line 15]. If it is performing self-evaluation it does not use exploration noise. 
            
            \item \textbf{LP update.} If the agent was performing self-evaluation, it can now update its measures of LP given the new result (success or failure, purple) [lines 17 and 18]. It computes the corresponding probability measures $p_{LP}$ using an $\epsilon$-greedy version of the proportional probability matching method such that:
                $$p_{LP}(M_i) = \epsilon \times \frac{1}{N} + (1-\epsilon) \times \frac{|LP_{M_i}|}{\sum_{j=1}^{N} |LP_{M_j}|}, $$
            where $LP_{M_i}$ is the learning progress of module $M_i$, and $\epsilon$ implements a mixture between random exploration (left-term) and exploitation guided by the absolute learning progress (right term).
            
            \item \textbf{Module and goal substitution.} The agent decides on which modules and goals to train. To update the policy and critic, the algorithm first samples a minibatch from the replay buffers (red) using $p_{LP}$, see \emph{Prioritized Interest Replay} below. If it wants to train on module $M_i$, it samples a transition relevant to module $M_i$ and substitutes the original module descriptor stored in the transition ($m_d$) by the module descriptor corresponding to $M_i$, ($m_d^*$ in the figure). With probability $p=0.8$, it also substitutes the current goal $g$ by an outcome achieved later in the episode using hindsight ($g^*$ in the figure). These substitutions enable cross-module and cross-goal learning [line 20].
            
            \item \textbf{Internal reward.} The agent computes its reward $r$ for each transition, using the internal reward function $R_{M,g}$ parameterized by the substitute module description ($m_d^*$) and the substitute goal ($g^*$) (brown) [line 21].
            
            \item \textbf{RL updates.} The agent updates its policy and value function using \ddpg (green) [line 22]. 
        \end{enumerate}

        \paragraph{Prioritized Interest Replay.} In \emph{Prioritized Experience Replay} (\textsc{PER}), \citet{schaul2015prioritized} suggests that RL agents can learn more efficiently from some transitions than from others. They propose to bias transition replay towards transitions with high temporal-difference (TD) error. In multi-goal settings, \citet{zhao2018energy} proposes to bias sampling towards episodes of higher energy, where energy is the sum of the potential energy, the kinetic energy and the rotational energy of the considered object (Energy-Based Prioritization or \textsc{EBP}). They show that \textsc{EBP} performs higher than \textsc{PER} in their experiments. Here, we propose to apply a simpler version of \textsc{EBP} that we call \textit{Prioritized Interest Replay}. To learn about module $M_i$ (e.g. \emph{Push}(cube 1) module), we only use transitions from episodes where the corresponding outcome (e.g. cube 1 position) has changed during the episode. This is an heuristic modeling attention towards \emph{interesting} episodes relative to the module at hand. Note that, contrary to \textsc{PER} where transitions are prioritized based on a criterion at the transition level (TD-error of the transition), we apply a criterion at the episode level (did the outcome change during the episode?), as in \textsc{EBP}. Although we pick interesting episodes, we still sample uniformly from those. We assume this heuristic to induce a small sampling bias and do not use any correction by importance sampling as it was done in \textsc{PER}. We apply that heuristic to all algorithms in our paper.
        
        After the interaction with the environment, transitions are stored in $N+1$ buffers. A transition is stored in buffer $i$ when the outcome corresponding to $M_i$ has changed during the episode. The additional buffer collects the remaining episodes. Before training, the agent decides how to allow its computational resources among the different modules using $p_{LP}$. If the size of the minibatch is $N_{mb}$, the agent will sample $\left \lfloor{N_{mb} \times p_{LP}(M_i)}\right \rfloor$ transitions from buffer $i$, this for all modules $M_i$. In this minibatch, every transition has been sampled to train on a specific module ($m_d^*$), although it could have been collected while targeting another module ($m_d$). To perform this cross-module learning, we simply substitute the module descriptor of each transition $m_d$ by $m_d^*$.
        
        \paragraph{Internal Rewards.} The internal reward function is parameterized by both the module and the goal $R_{M,g}$. It is positive ($r=0$) when the constraints defined by the modules are satisfied for the current goal and outcome. For the \emph{Push} and \emph{Pick and Place} modules, the reward is positive  when the distance between the cube and the target position is within $\epsilon_{reach} = 0.05$ (simulation units), negative otherwise. For the \emph{Reach} module, the same criterion is applied to the distance between the gripper and the goal. Finally, for the Stack module, in addition to the constraints on the goal-cube distance, the goal-gripper distance must stay larger than $1.2 \times \epsilon_{reach}$, to ensure that the two cubes are actually stacked and not hold by the gripper. With this example we see that a module simply defines a set of constraints. Although here it is applied to a single state, we could imagine constraints defined over multiple time-steps (e.g. keep the cube above the table for more than 2 seconds).

    \subsection*{Environment}
        \emph{Modular Goal Fetch Arm} is a new simulated environment based on the Fetch environments included in the OpenAI Gym suite \citep{brockman2016openai}. The agent is embodied by a 7-DoF Fetch robotic arm facing 2 cubes randomly positioned on a table and can target a large set of diverse goals. Additional out-of-reach cubes can be added as distracting modules. The agent controls the $3$D Cartesian position of its gripper in velocity as well as its two-fingered parallel gripper. The agent can target several modules of goals: ($M_1$) reaching a $3$D target with the gripper; ($M_2$) reaching a $2$D target on the table with cube $1$; ($M_3$) reaching a $3$D target with cube $1$; ($M_4$) stacking cube $1$ over cube $2$; ($M_{5-}$) reaching a $2$D target on the table with one of the randomly moving cube which are out of reach (distracting module). The observation space has 40 dimensions (see original paper presenting the Fetch environments for details \cite{plappert2018multi}) + 3 per additional out-of-reach cube, while the action space has 4 ($3$D actions + gripper).

    \subsection*{Evaluation Methodology}
        Every epoch ($50\times19$ actors $=950$ episodes), the $19$ actors are evaluated offline on $5$ rollouts each. For each evaluation rollout, the experimenter asks the agent to perform a goal selected at random in the set of achievable goals (subset of goals the agent trained on). This external evaluation of the agent's true competence is distinct from the self-evaluation performed by the agent during learning. Note that the agent policy is frozen during evaluation, and that the agent cannot use the evaluation trajectories for later training as opposed to trajectories obtained by self-evaluation. Since all actors share the same policy, each point $\hat{p}$ of a learning curve represents the maximum likelihood estimate of a Bernoulli probability $p$ (i.e. success rate) using a sample size of $n=19 \times 5 = 95$ rollouts. With this sample size, the confidence interval can be estimated using a normal distribution:
        $$error=|p-\hat{p}| \leq 1.96 \sqrt{\frac{\hat{p}(1-\hat{p})}{n}},$$ 
        with probability $95\%$, which is itself upper-bounded by $1.96\sqrt{\frac{0.25}{95}}=0.1$.
    
        Each experiment is replicated with $10$ different random seeds. Because the sample size is relatively small ($<30$), we cannot make any assumption about the distribution of the performance measures. Comparisons are therefore conducted using a one-tail Mann-Whitney U-test, a non-parametric test that does not make such assumptions \cite{mann1947test}. We use a type-I error (confidence level) $\alpha=0.01$. The one-tail version is used because we expect our algorithm to perform better.

    \subsection*{Hyperparameters}
        The \curious algorithm is built on top of the OpenAI Baselines implementation of \her-\ddpg.\footnote{ The OpenAI Baselines implementation of \her-\ddpg can be found at https://github.com/openai/baselines, our implementation of \curious can be found at \url{https://github.com/flowersteam/curious}.} This consists in a parallel implementation with $19$ actors. The actors share the same parameters and their updates are averaged to compute the next set of parameters. We use the same hyperparameters as \citet{plappert2018multi}. Note that this paper does not focus on the underlying learning algorithm, which could be replaced by any off-policy reinforcement learning algorithm. 
    
        \curious uses three extra hyperparameters. 1) The probability $p_{eval\_c}$ that the agent performs self-evaluation for the next rollout. It does so to update its subjective competence measure for a module, $p_{eval\_c}=0.1$. 2) The length of the windows considered to compute the subjective measures of competence and learning progress is set to $l=300$ episodes. 3) The exploration parameter $\epsilon$ which controls the mixture between random module selection and active module selection based on absolute learning progress is set to $\epsilon=0.4$.

    \section*{Additional Results}
    
    \subsection*{Developmental Learning: Regularities and Diversity}
        \label{sec:developmental}
        Traditionally, the study of learning phases is based on behavior. This is true in Psychology, which can only access the human behaviors but cannot access internal representations. Multi-goal RL papers as well, when they do provide behavioral studies, usually present the test performance of the agent on various goals as a function of time (e.g. \cite{mankowitz2018unicorn}). In the present paper however, we have a direct access to the internal representations of our agents, to their beliefs about their current competence and learning progress for each of the modules. It is from these internal representations that we study developmental progressions. This enables us to distinguish active curriculum learning from passive experience of the environmental structure. Indeed, the structure of the environment can be such that some modules are simpler than others, leading random agents to learn them faster than others. On the other hand, agents performing active module selection such as \curious, actively guide their learning trajectories as a function of their internal representations of competence and progress.
        
\begin{figure*}\sffamily
            \vskip 0.15in
            \begin{center}
            \begin{small}
            \begin{tabular}{ccc}
            
                \toprule
                \textsc{Trials} & \textsc{Competence} & \textsc{Learning Progress} \\
                \midrule
                1 &  \includegraphics[width=0.3\textwidth]{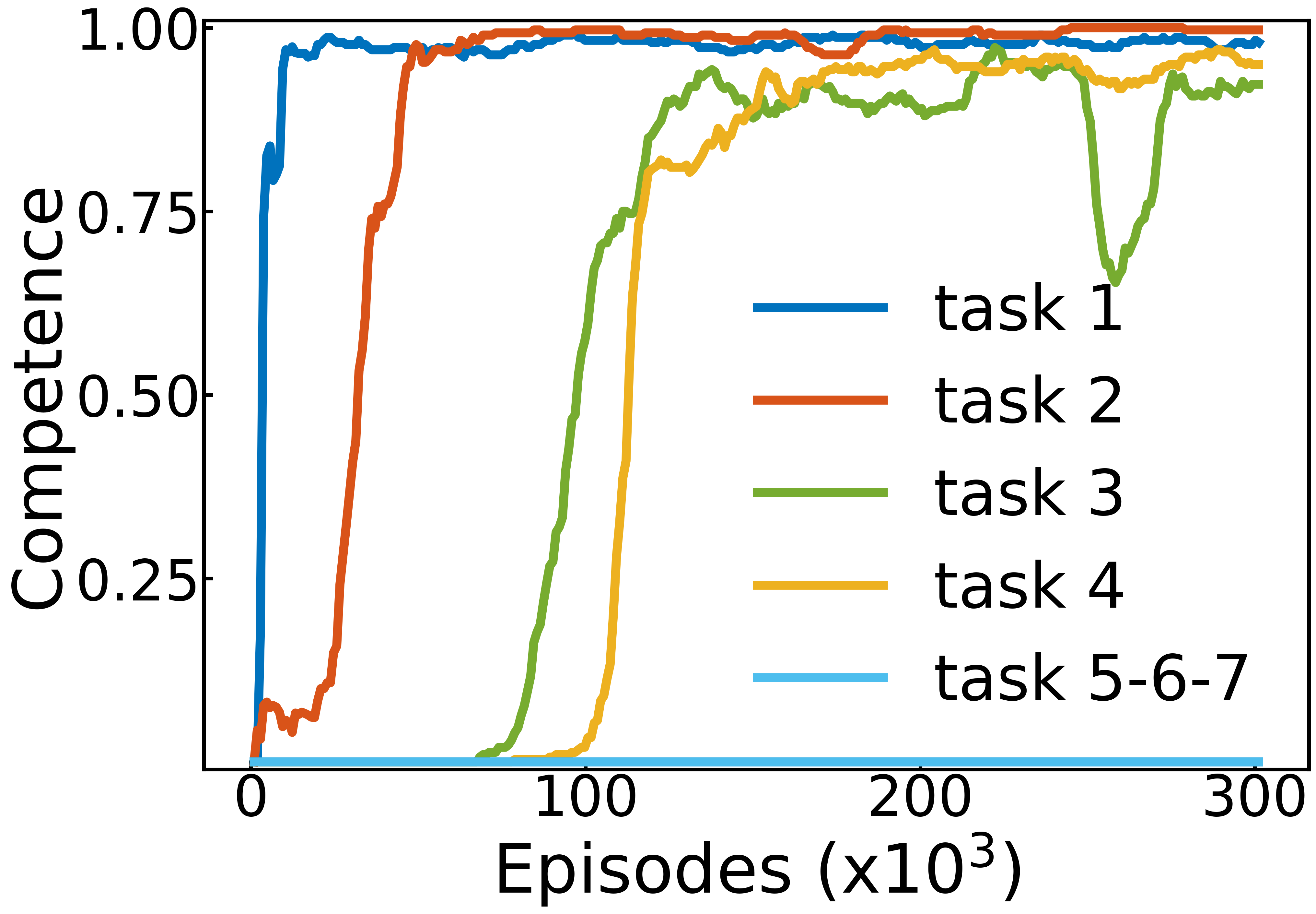} & \includegraphics[width=0.3\textwidth]{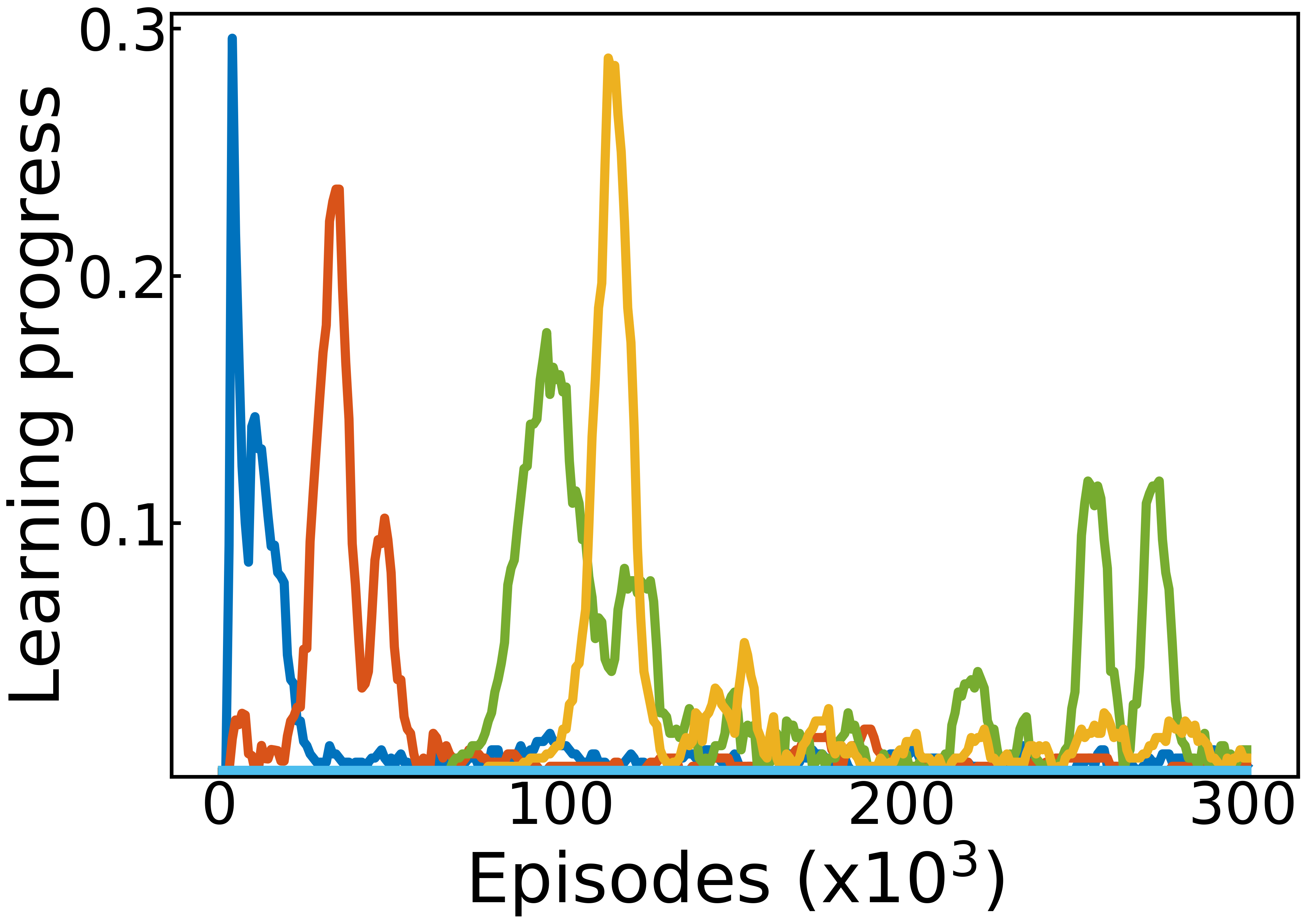} \\
                2 & \includegraphics[width=0.3\textwidth]{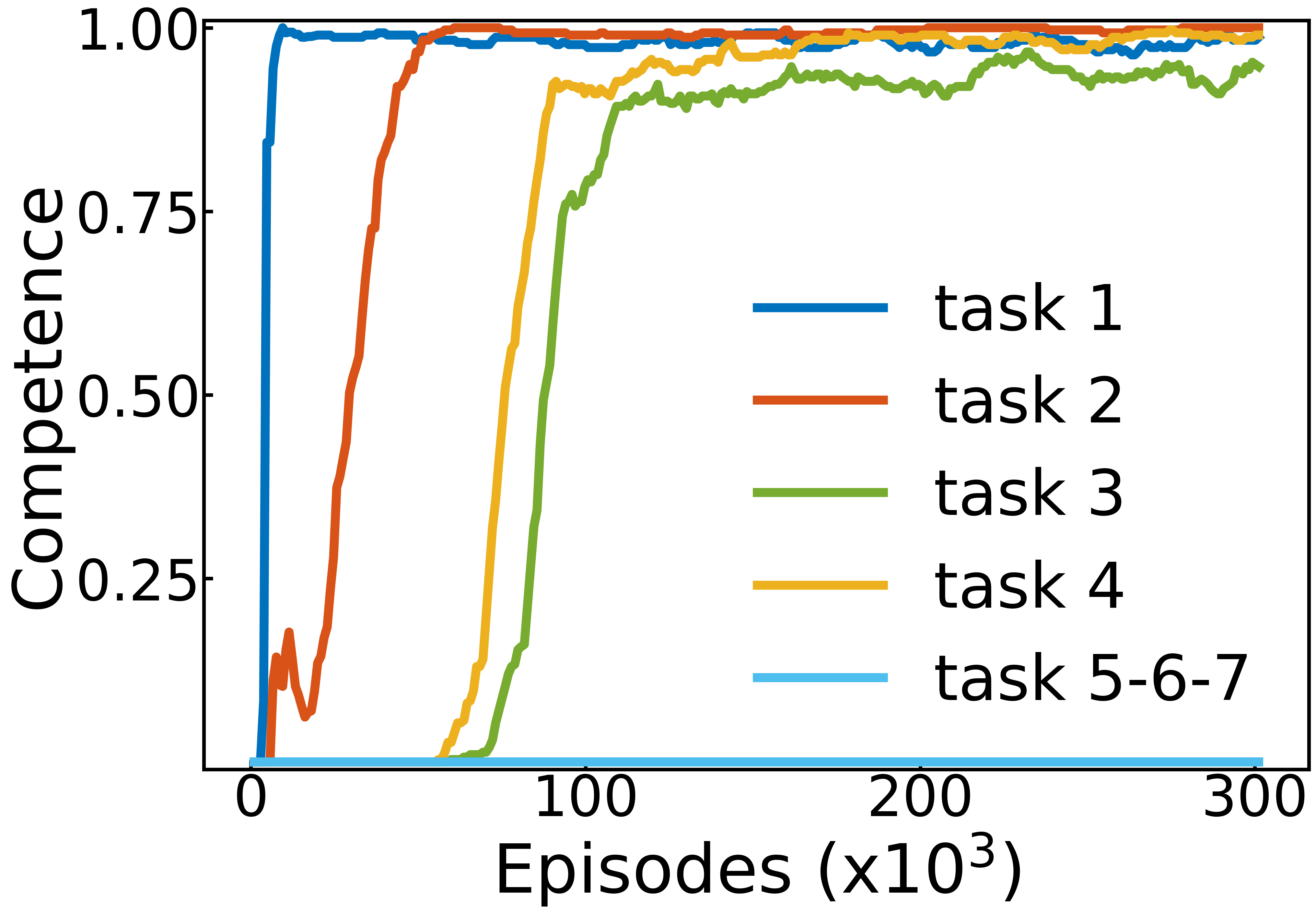} & \includegraphics[width=0.3\textwidth]{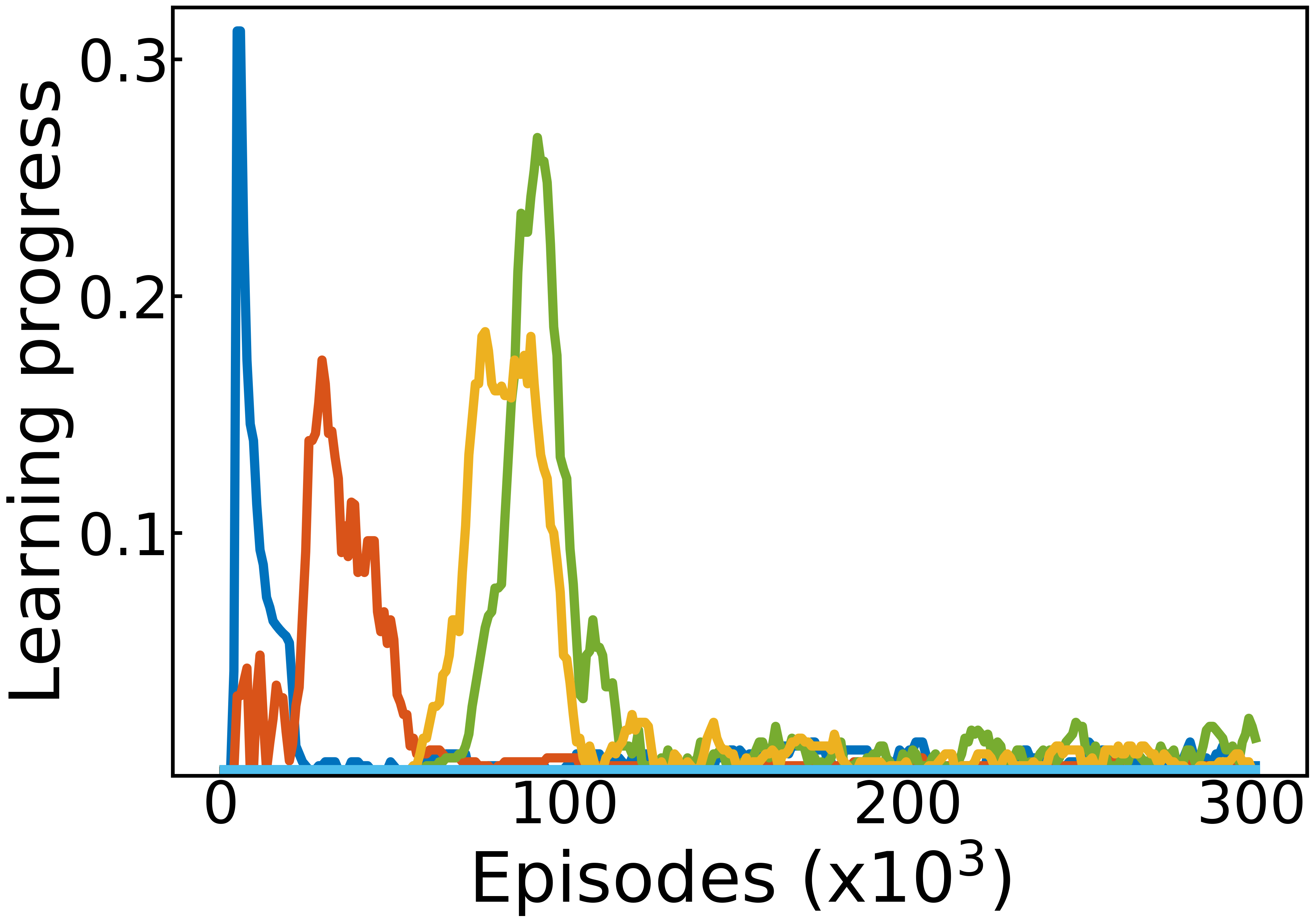} \\
                3 & \includegraphics[width=0.3\textwidth]{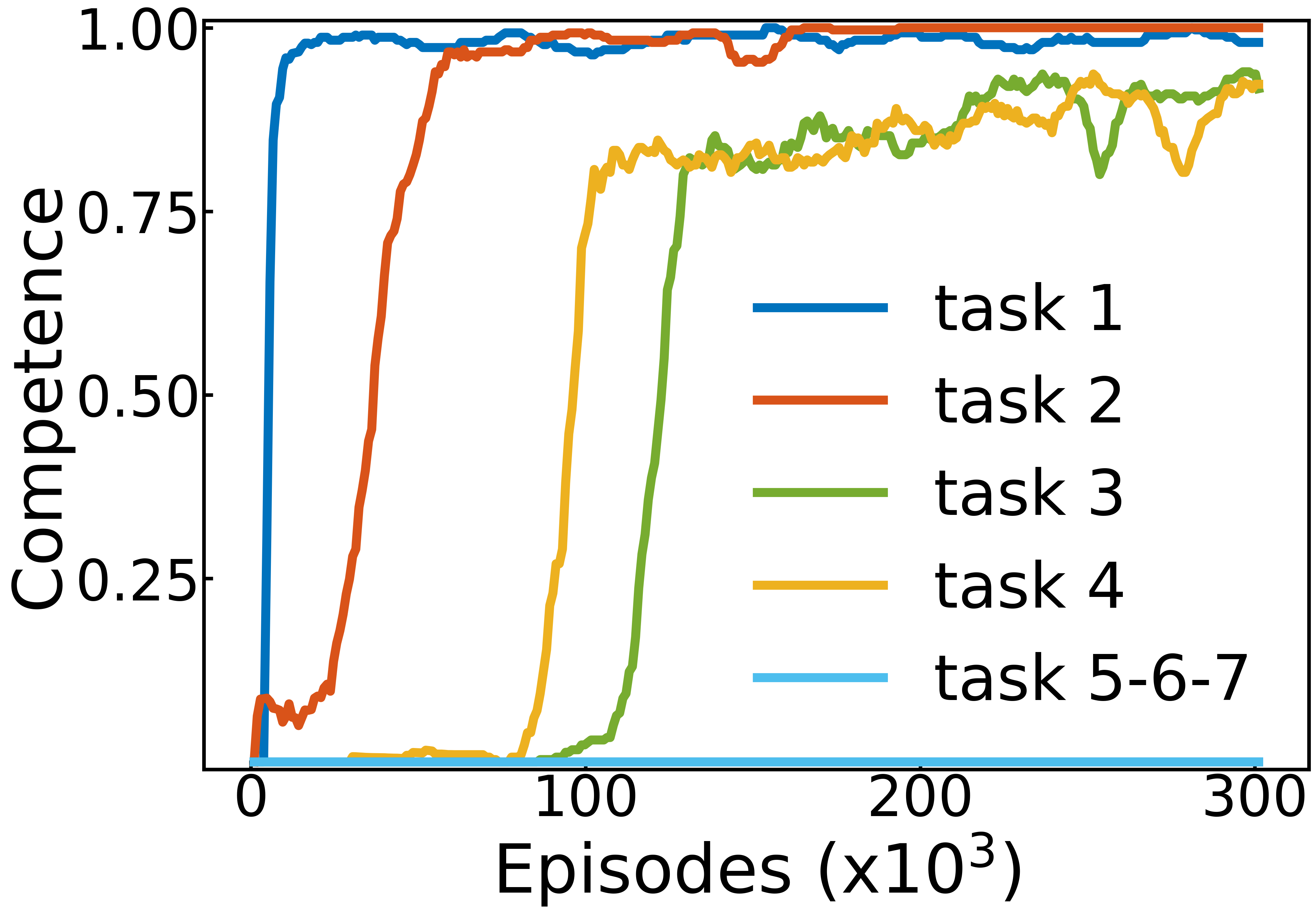} & \includegraphics[width=0.3\textwidth]{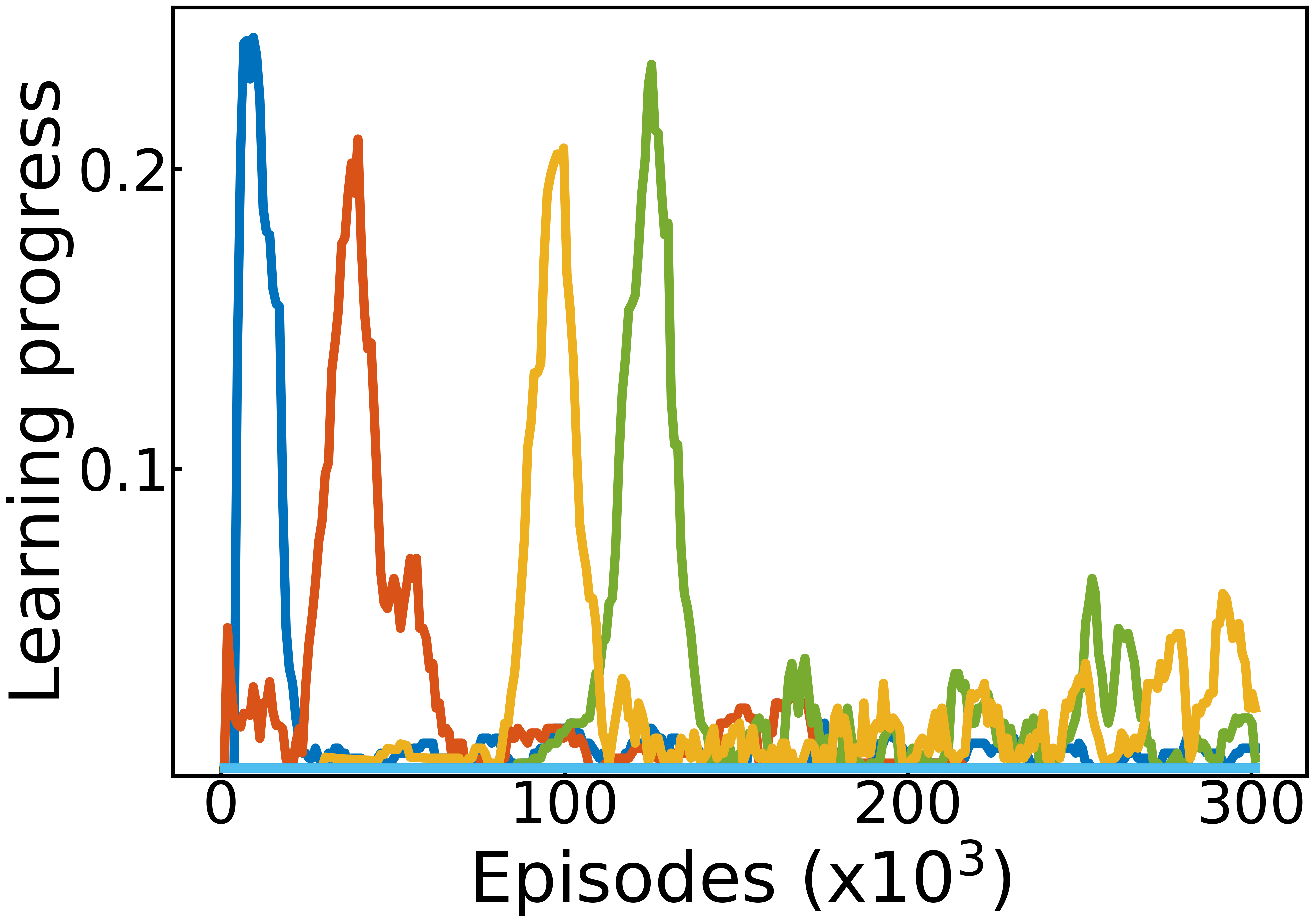} \\
                4 & \includegraphics[width=0.3\textwidth]{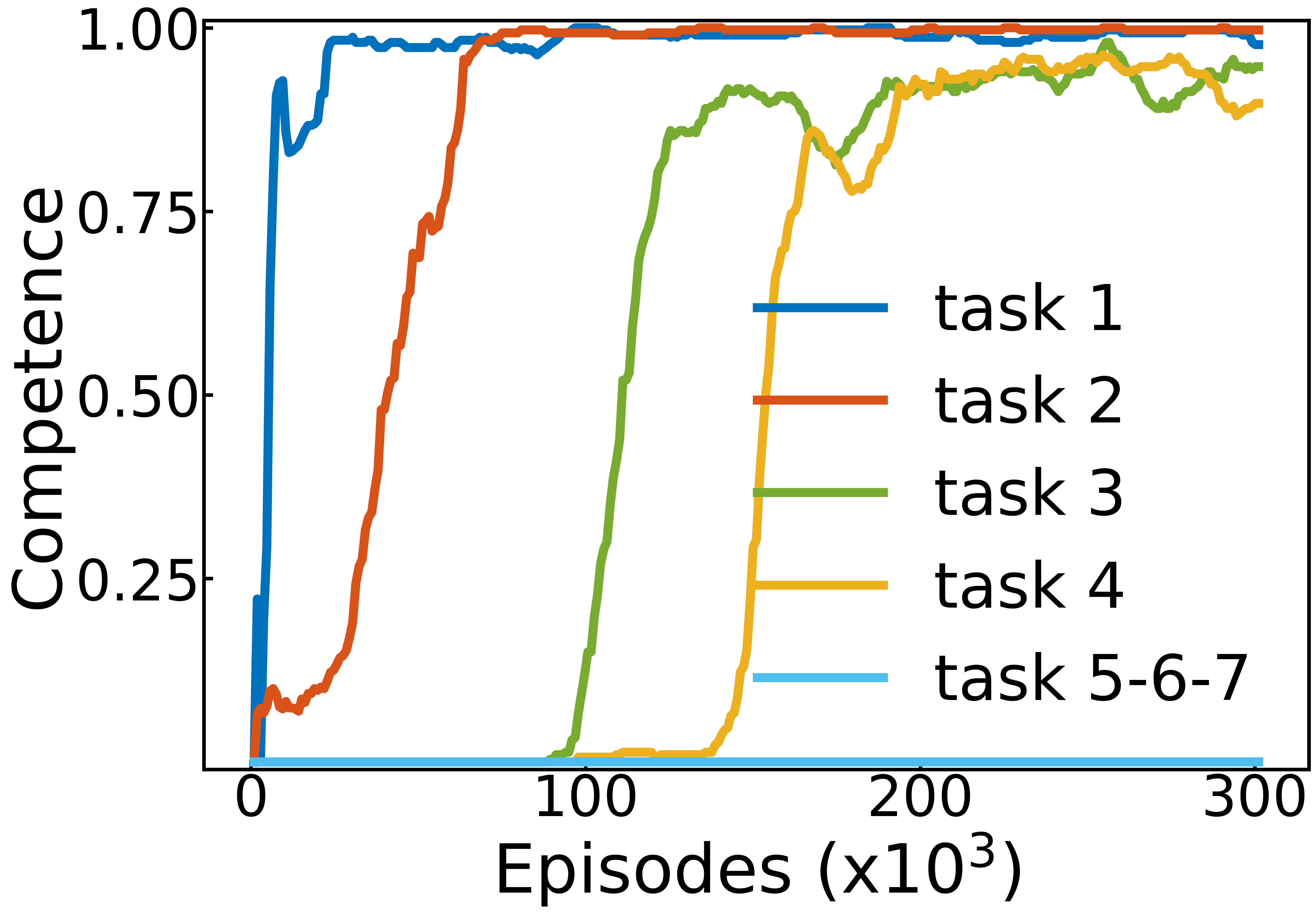} & \includegraphics[width=0.3\textwidth]{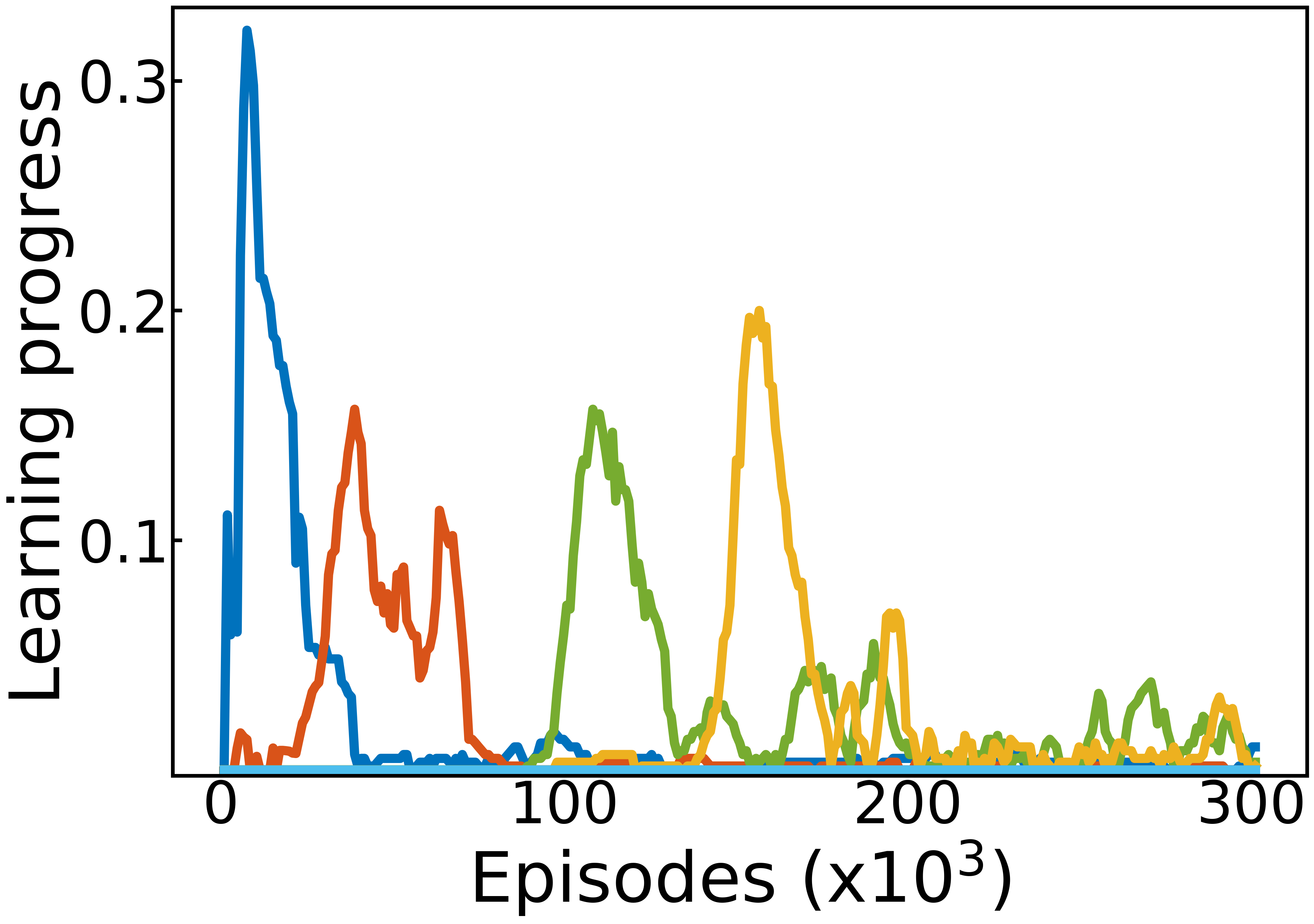} \\
                5 & \includegraphics[width=0.3\textwidth]{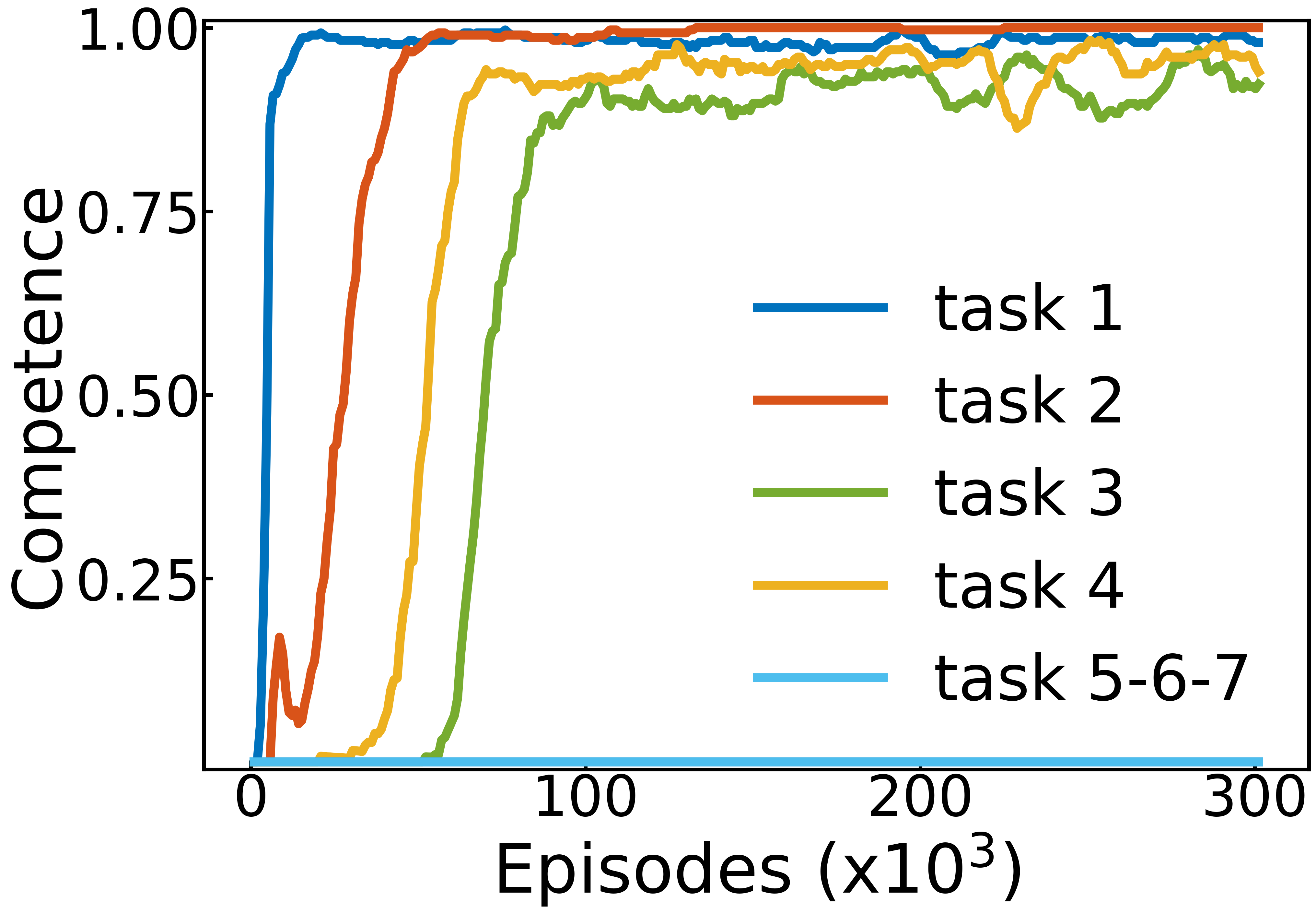} & \includegraphics[width=0.3\textwidth]{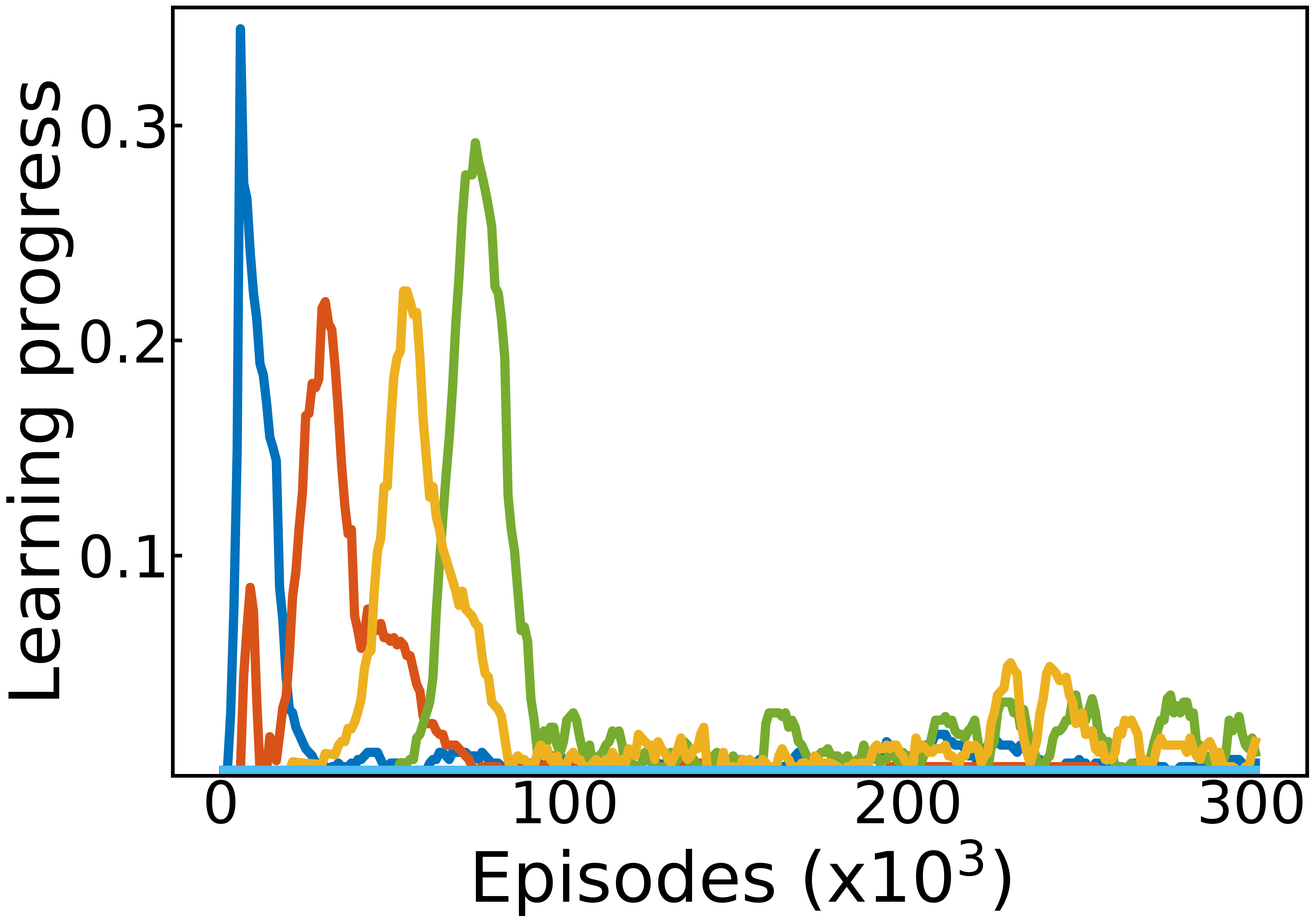} \\
            \end{tabular}
            \end{small}
            \end{center}
            \vskip -0.1in
            \caption{\textbf{Learning Phases}. Competence (left) and absolute learning progress (right) for 5 trials of the \curious algorithm (Reach, Push, Pick and Place, Stack + 3 distracting modules).}
            \label{fig:all_c}
        \end{figure*}

        In Fig.~\ref{fig:all_c}, we present 5 sets of subjective competence curves and their associated learning progress. We can see that $M_1$ (\emph{Reach} module) is always learned first and $M_2$ (\emph{Push} module) is always learned second. However, $M_3$ (\emph{Pick and Place}) and $M_4$ (\emph{Stack}) can be learned in various order or even simultaneously depending on the individual learning trajectories. Once the agent has experienced a few successes on a module, LP increases and the agent focuses on it, which generates even more successes. What happened by chance in the early stages of learning leads this agent to focus first on either $M_3$ or $M_4$. Although some modules might be easier to learn first, or necessary to learn others, individual experience can influence learning trajectories just as for humans \citep{oudeyer2016evolution}.
    
    \subsection*{Scaling Properties}
    It is important to discuss the ability of \curious to scale to larger sets of modules. Quick tests showed that \curious could scale to at least 10 achievable modules (Reach, Push, Pick and Place and Stack with different cubes). However, we can expect \curious to fail when the number of modules gets even larger. This could be mitigated by enabling \curious to deal with multiple actors and critics, each pair dealing with a subset of the modules.

    \subsection*{Meaning of the Term \emph{Goal}}
        It is important to note that the term \emph{goal} used in the context of Intrinsically Motivated Goal Exploration Processes (\imgep) is much more general than the one used in this paper. Indeed, in \imgep, \emph{self-generated goal} denotes any self-defined parameterized problem, which solution should be found through one’s own actions (e.g. it can as diverse as ``Grasp \emph{obj1} and place it at \emph{pos3}'', ``Move to \emph{(x,y,z)}'', ``Find a \emph{blue} key'', ``Collect an even number of \emph{obj3}'', or "generate a trajectory that contains 3 loops"). 
    
    \subsection*{Computational Resources}
        One trial of one algorithm takes around 20 hours to run on $19$ cpus. This paper contains around $130$ trials, which sums to a total of $130 \times 20 \times 19 \approx 5.6$ cpu years.

\end{document}